\documentclass[letterpaper, 10 pt, journal, twoside]{IEEEtran}

\usepackage{generic}
\usepackage{color}
\usepackage{cite}
\usepackage{amsmath,amssymb,amsfonts}
\usepackage{algorithmic}
\usepackage{graphicx}
\usepackage{textcomp}
\usepackage{subfigure}
\usepackage{ulem}
\usepackage{caption}
\usepackage{float} 

\newcommand{\eproof}{\hfill\rule{1mm}{1mm}}

\newcommand{\bstate}{\medskip\begin{state} }
	\newcommand{\estate}{ \hfill  \rule{1mm}{2mm}\medskip\end{state}}

\newcommand{\bass}{\medskip\begin{ass} }
	\newcommand{\eass}{ \hfill  \rule{1mm}{2mm}\medskip\end{ass}}

\newcommand{\brem}{\medskip \begin{remark}  }
	\newcommand{\erem}{\hfill \rule{1mm}{2mm}\medskip
\end{remark} }
\newcommand{\bthm}{\medskip\begin{theorem}  }
	\newcommand{\ethm}{ \hfill  \rule{1mm}{2mm} \medskip
\end{theorem} }
\newcommand{\blem}{\medskip\begin{lemma}  }
	\newcommand{\elem}{ \hfill \rule{1mm}{2mm}\medskip
\end{lemma} }
\newcommand{\bcorollary}{\medskip\begin{corollary}  }
	\newcommand{\ecorollary}{  \hfill \rule{1mm}{2mm}\medskip
\end{corollary} }
\newcommand{\bdefn}{\medskip\begin{definition}}
	\newcommand{\edefn}{  \hfill \rule{1mm}{2mm}\medskip
\end{definition} }
\newcommand{\bproposition}{\medskip\begin{proposition} }
	\newcommand{\eproposition}{\hfill \rule{1mm}{2mm}\medskip
\end{proposition} }
\newcommand{\bexample}{\medskip\begin{example} \rm}
	\newcommand{\eexample}{ \hfill \rule{1mm}{2mm}\medskip
\end{example} }

\newcommand{\bcon}{\medskip\begin{condition} \rm}
	\newcommand{\econ}{ \hfill \rule{1mm}{2mm}\medskip
\end{condition} }

\renewcommand{\t}{^{\top}}
\newcommand{\proofnow}{\noindent{\bf Proof: }}

\newtheorem{theorem}{\bf Theorem}[section]
\newtheorem{ass}{\bf Assumption}[section]
\newtheorem{lemma}{\bf Lemma}[section]
\newtheorem{definition}{\bf Definition}[section]
\newtheorem{remark}{\bf Remark}[section]
\newtheorem{corollary}{\bf Corollary}[section]
\newtheorem{proposition}{\bf Proposition}[section]
\newtheorem{example}{\bf Example}[section]
\newtheorem{condition}{\bf Condition}[section]
\newtheorem{state}{\bf Assumption}[section]

\begin{document}
\title{A Whole-Body Disturbance
	Rejection Control Framework for Dynamic Motions in 	Legged Robots \vspace {-0.2em}
\author{Bolin Li$^{1}$, Wentao Zhang$^{1}$, Xuecong Huang$^{1}$, Lijun Zhu$^{2}$, and Han Ding$^{3}$}
\thanks{Received 17 June 2025; accepted 5 July 2025. Date of publication; date of current version. This article was recommended for publication by Associate Editor K.Akbari Hamed and Editor A.Kheddar upon evaluation of the reviewers’ comments. This work was supported in part by the National Natural Science Foundation of China under Grant 62173155 and Grant 52188102 and in part by the Taihu Lake Innovation Fund for Future Technology, HUST. \textit{(Corresponding author: Lijun Zhu)}} 
\thanks{$^{1}$Bolin Li, Wentao Zhang, and Xuecong Huang are with the School of Artificial Intelligence and Automation, Huazhong University of Science and Technology, Wuhan 430074, China {\tt\footnotesize bolin\_li@hust.edu.cn, wentaozhang@hust.edu.cn, M202373511@hust.edu.cn}}
\thanks{$^{2}$ Lijun Zhu is with the School of Artificial Intelligence and Automation,
	Huazhong University of Science and Technology, Wuhan 430074, China, and
	also with the State Key Laboratory of Intelligent Manufacturing Equipment and
	Technology, Huazhong University of Science and Technology, Wuhan 430074,
	China {\tt\footnotesize ljzhu@hust.edu.cn}}
\thanks{$^{3}$Han Ding is with the State Key Laboratory of Intelligent Manufacturing
	Equipment and Technology, Huazhong University of Science and Technology,	Wuhan 430074, China, and also with the School of Mechanical Science and Engineer, and the State Key Laboratory of Intelligent Manufacturing Equipment and Technology, Huazhong University of Science and Technology, Wuhan	430074, China
	{\tt\footnotesize dinghan@hust.edu.cn}}
	\thanks{This article has supplementary downloadable material available at https://doi.org/10.1109/LRA.2025.3588716, provided by the authors.}
\thanks{Digital Object Identifier 10.1109/LRA.2025.3588716}
	}

%\thanks{S. B. Author, Jr., was with Rice University, Houston, TX 77005 USA. He is 
%now with the Department of Physics, Colorado State University, Fort Collins, 
%CO 80523 USA (e-mail: author@lamar.colostate.edu).}
%\thanks{T. C. Author is with 
%the Electrical Engineering Department, University of Colorado, Boulder, CO 
%80309 USA, on leave from the National Research Institute for Metals, 
%Tsukuba, Japan (e-mail: author@nrim.go.jp).}
\maketitle
\begin{abstract}
This letter presents a control framework for legged robots that enables self-perception and resistance to external disturbances and model uncertainties. First, a novel disturbance estimator is proposed, integrating adaptive control and extended state observers (ESO) to estimate external disturbances and model uncertainties. This estimator is embedded within the whole-body control framework to compensate for disturbances in the legged system. Second, a comprehensive whole-body disturbance rejection control framework (WB-DRC) is introduced, accounting for the robot's full-body dynamics. Compared to previous whole-body control frameworks, WB-DRC effectively handles external disturbances and model uncertainties, with the potential to adapt to complex terrain. Third, simulations of both biped and quadruped robots are conducted in the Gazebo simulator to demonstrate the effectiveness and versatility of WB-DRC. Finally, extensive experimental trials on the quadruped robot validate the robustness and stability of the robot system using WB-DRC under various disturbance conditions. 
\end{abstract}


\section{Introduction}
\IEEEPARstart{L}{egged} robots have gained significant attention in recent years due to their ability to navigate complex and uneven terrain \cite{9966331}, offering substantial advantages in a wide range of applications, such as search and rescue, exploration, and assistive robotics. However, one of the main challenges in legged robot locomotion is the ability to maintain stability and robustness in the face of external disturbances and model uncertainty, such as ground irregularities, unexpected external forces, heavy loads,  or changes in terrain conditions. These disturbances can significantly affect the robot's state and lead to deviations from the desired trajectory, ultimately compromising the performance and stability.

To enhance the robustness of legged robots, various methods for disturbance rejection have been developed and implemented. One approach involves the design of sophisticated controllers, such as limit cycle walkers \cite{4456910, 10335933} and high-performance optimal control \cite{10214438,sombolestan2024adaptive}, to improve disturbance rejection. The article   \cite{sombolestan2024adaptive} presents a novel adaptive control methodology integrated into a force-based control system, enabling agile-legged robots to carry heavy loads while maintaining dynamic motion over rough terrain, as demonstrated experimentally on the Unitree A1 robot. Another approach utilizes disturbance rejection observers \cite{zhu2023proprioceptive} to further enhance the robot's robustness. The general momentum observer is a useful method for the disturbance perception. Initially developed for collision detection and disturbance perception in robotic manipulator arm systems \cite{6290917}, this approach has been widely adopted in recent years for legged systems \cite{ 8460904, morlando2021whole, zhu2023proprioceptive}. In \cite{zhu2023proprioceptive}, a proprioceptive-based disturbance estimator is proposed, which effectively filters foot-ground interaction noise and suppresses estimation errors. It is paired with a hierarchical whole-body controller that accounts for full-body dynamics, actuation limits, external disturbances, and interactive constraints.
In \cite{8460904}, a discrete-time extension of the generalized-momentum disturbance observer is presented to enhance the accuracy of proprioceptive force control estimates, with the information fused with other contact priors under a Kalman filtering framework to improve the robustness of the method. In \cite{morlando2021whole}, an external disturbance estimator was proposed for legged robots based on the system's momentum, integrating a motion planner, an optimization problem for ground reaction forces, and a whole-body controller, which is tested on a quadruped robot in a dynamic simulation. These generalized-momentum disturbance observer rely on the assumption that the relationship between the inertia matrix and the Coriolis matrix of the Euler-Lagrange system holds, as discussed in \cite{8460904}, which limits their applicability.  Although disturbance observers are effective for external disturbance rejection and compensation in legged robots, they encounter difficulties in handling model uncertainties and fault tolerance, especially in robots with complex dynamic interactions and inaccurate motor output torque. Previous model-based control methods for addressing uncertainties in legged robots either overlooked high-level planning and model uncertainties \cite{zhu2023proprioceptive}, failed to account for external disturbances and fault tolerance \cite{10214438, 10705076}, or failed to incorporate fault tolerance and achieve performance based on a simplified rigid body model \cite{sombolestan2024adaptive}. Additionally, these approaches in \cite{10214438, 10705076, sombolestan2024adaptive} were based on single rigid-body dynamics and primarily applied to quadruped robots, making them less suitable for bipedal and humanoid robots.

An active disturbance rejection control strategy (ADRC)  was developed in \cite{4796887} to address plants with significant uncertainties in both dynamics and external disturbances \cite{ran2021new, 10226466}. The core design of ADRC involves using an ESO to estimate generalized disturbances and compensate for them in a feedforward manner. Subsequently, a simple proportional-derivative control law is synthesized to ensure bounded-error tracking performance. While ADRC is effective in handling various modeling uncertainties when they are accurately estimated by the ESO, it does not explicitly account for parameterizable uncertainties, which increases the learning burden on the ESO. As a result, compared to adaptive control, ADRC tends to exhibit poorer tracking performance in systems with substantial parametric uncertainties \cite{7900325}.  The adaptive control approach is typically used to address parametric uncertainties within a single controller \cite{sombolestan2024adaptive, 10705076}. By integrating adaptive control and ESO through the full-state feedback, where parametric uncertainties and uncertain nonlinearities are handled separately, improved tracking performance can be achieved, as verified in hydraulic servo systems \cite{7900325}.

Unlike previous work, we develop a novel disturbance estimator that addresses model uncertainty, external disturbances, and fault tolerance, integrating it into the control architecture of legged robots. Our approach combines the ESO and adaptive control to estimate disturbances while considering whole-body uncertainties. Inspired by the work in \cite{sombolestan2024adaptive}, we adopt a new form for model uncertainty, incorporating unknown parameters and nonlinearities, rather than estimating a total disturbance. This form offers better physical interpretability and enhances disturbance rejection performance. The main contributions and important emphases are the following: 

1)  A novel disturbance estimator has been introduced, integrating adaptive control and an ESO to estimate uncertainties in the legged system. The stability and boundedness of the   proposed estimator are guaranteed  with or without  uncertainties.

2) A general whole-body disturbance rejection control framework is proposed and is integrated into a hierarchical    control architecture of legged robots.

3) Simulations are conducted on both biped and quadruped robots, and experiments are carried out on the Unitree A1 quadruped robot to verify the effectiveness of the proposed WB-DRC, as well as its advantages in terms of robustness and stability.

This letter is organized as follows. Section \ref{sec:Preliminaries} provides the background on legged robots. Section \ref{sec:hlc} presents the high-level control.
Section \ref{sec:disturbance_estimator} introduces the adaptive extended state disturbance rejection estimator. Section \ref{sec:disturbance_rejction_WBC} presents the whole-body disturbance rejection planner. Sections \ref{sec:simulation} and \ref{sec:experiment} evaluate the performance of the proposed framework through simulation and on a physical prototype, respectively. Finally, Section \ref{sec:conclusions} concludes the letter.

\section{Preliminaries}\label{sec:Preliminaries}
The whole-body dynamics of a floating-base robot  can be written as
\begin{gather}
	D(q)\ddot{q} + C(\dot{q},q)\dot{q} + G(q) = S\t \tau + J(q)\t F + d \label{eq:dynamics}
\end{gather}
where  $q:=[(q^b)\t,(q^j)\t]\t\in\mathbb{R}^{6+n}$ with
 $q^{b} \in \mathbb{R}^6$ being the pose of the floating base  and $q^{j}$ being the actuated joint positions. 
 $d \in \mathbb{R}^{6+n}$ is the uncertainty,
 $D(q) \in \mathbb{R}^{(6+n)\times(6+n)}$ is a positive definite inertia matrix, $C(\dot{q}, q) \in \mathbb{R}^{(6+n)\times(6+n)}$ is the Coriolis matrix, $G(q) \in \mathbb{R}^{6+n}$ is the gravity vector;  
$S = [0_{n \times 6}, I_n]$ is  the selection matrix of the actuation; $\tau \in \mathbb{R}^{n}$ is the actuation torques; $F \in \mathbb{R}^{3 n_c}$ is the ground reaction forces (GRFs) of the stance leg; $J(q)\t F \in \mathbb{R}^{(6 + n) \times 3n_c}$ represents {the
	ground reaction forces} projected to the joints space, where $J(q)$
	is the contact-dependent Jacobian matrix; 
$n$ and $n_c$ 
represent the number of active degrees of freedom and the number of legs in the support surface, respectively.  For a humanoid robot  with rectangular feet, the contact points correspond to the four vertices of the rectangular shape \cite{8558661}.

For the high-level planning, we will use centroidal dynamics that describes the change of the linear and angular momentum with respect to the external wrench.  Denote by  $c \in \mathbb{R}^{3}$ the position of the COM in the inertial frame,   $h$ the linear momentum and ${L}$ the angular momentum around the COM of the robot. The linear momentum is  $h = m\dot{c}$ with $m$ being the total mass of the robot. 
The centroidal dynamics \cite{orin2008centroidal} can be  obtained according to the Newton-Euler law as:
\begin{align}
	\dot h  =& \sum\limits_{i = 1}^{{n_c}} {{f_i}}  - mg\nonumber \\
	\dot {{L}}  =& \sum\limits_{i = 1}^{{n_c}} {({p_i} - c) \times {f_i}} \label{eq:centroidal_dynamics}
\end{align}
where {$f_i \in \mathbb{R}^3$ represents the ground reaction force of contact $i$,} $p_i$ denotes the position of contact $i$, and $g = [0\;\;0\;\;9.81]\t$ is the gravity vector.

\begin{figure}[!ht]
	\centering
	\includegraphics[scale=0.52]{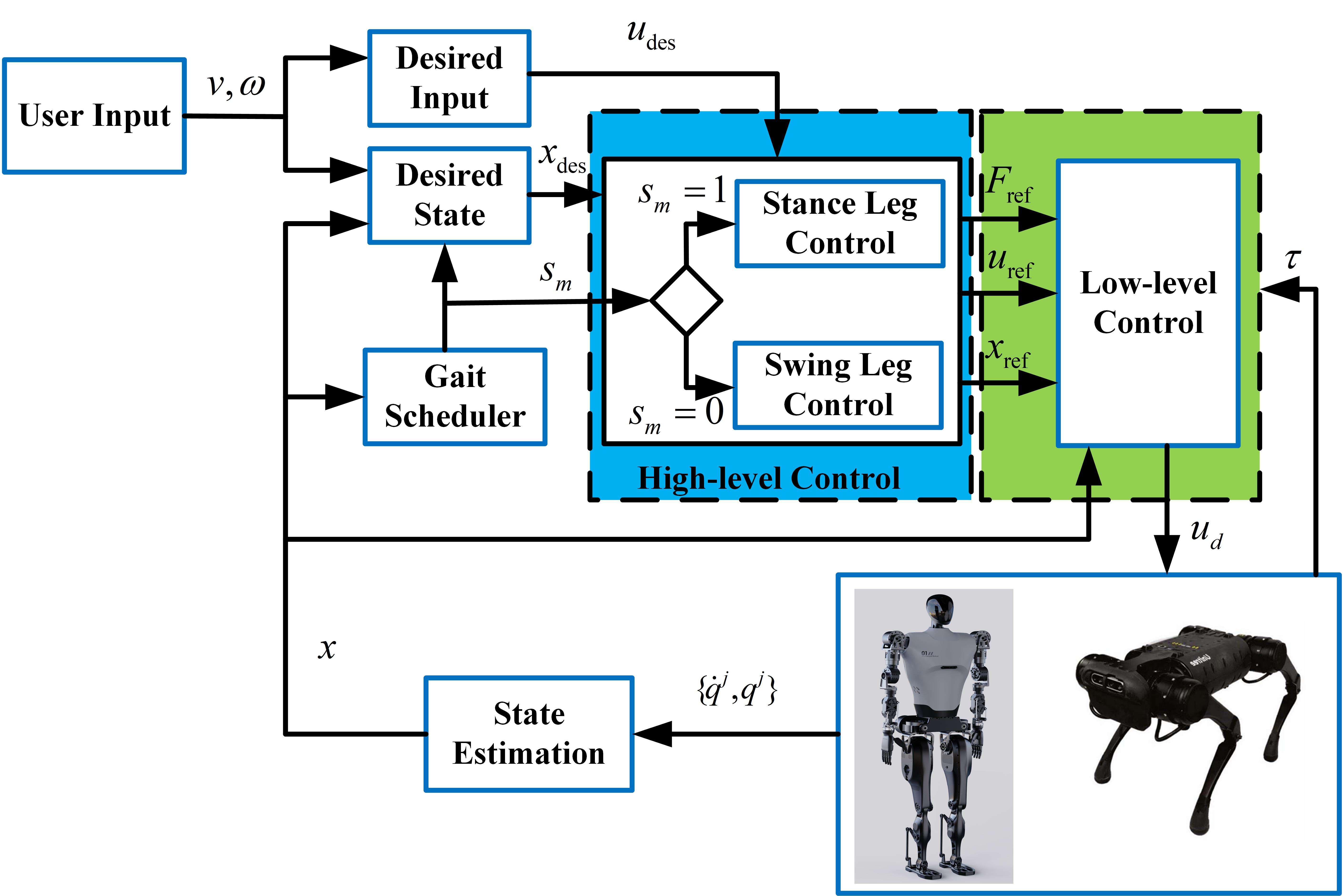}
	\caption{Block diagram of a control architecture
		for legged robots.}
	\label{fig_control_structure}
\end{figure}

In this letter, we adopt  the hierarchical control
architecture  used in {\cite{8593885}}.
%As in \cite{8593885},
As  in Fig. \ref{fig_control_structure},
 the robot’s control system consists of several modules, including a high-level controller, low-level controller, state estimation, and gait scheduler. The user input module receives the robot's  speed command  from the user input, and the gait scheduler module determines the contact times and sequences with $s_m \in \{1 =\text{contact}$, $0 = \text{swing}\}$. 
 The high-level control module employs the model predictive control (MPC) based on the centroidal dynamics to generate reference trajectories for the system state and inputs of the whole-body dynamics, which are then provided to the low-level leg control module.
% The high-level control module  will be detailed  in Section \ref{se:centroidal_dynamics}.
The low-level control module receives the reference trajectories and implements
the whole-body control (see \cite{grandia2023perceptive}) or inverse dynamics to maintain the balanced locomotion.

\begin{figure}[!ht]
	\centering
	\includegraphics[scale=0.48]{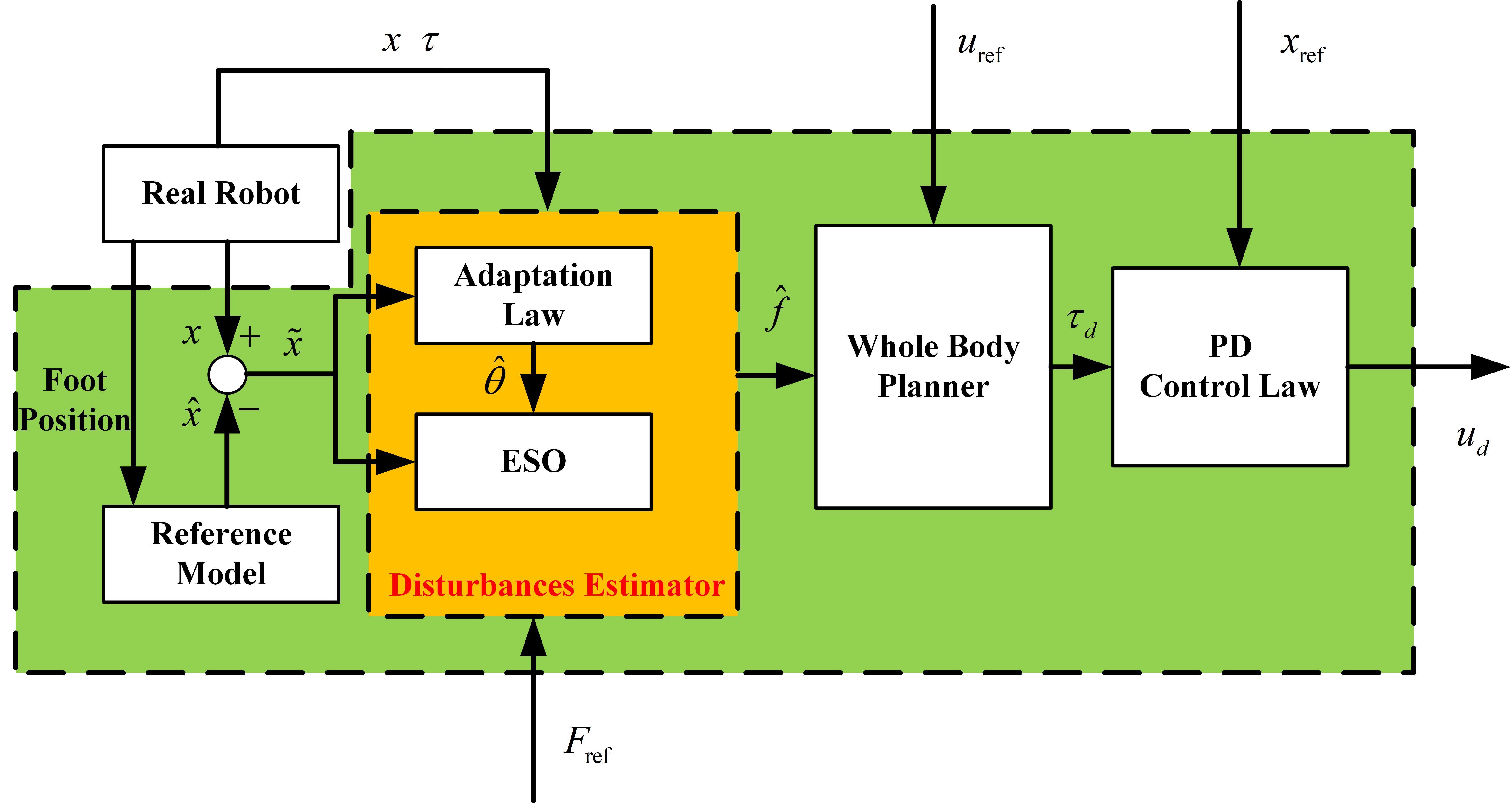}
	\caption{Block diagram of the WB-DRC.}
	\label{fig_disturbance_rejection}
\end{figure}

%%{\blue 
%Due to the unmodel dynamics and unknown external disturbance, the robust 	
%Legged locomotion must consider the disturbance rejection capability.
%
% should be capable of carrying heavy loads, possess strong fault tolerance, and effectively handle external disturbances. Achieving good performance in these practical requirements may not be possible without properly accounting for disturbances.}
To handle the unmodel dynamics and unknown external disturbance for the legged locomotion,
 a novel disturbance rejection whole-body control scheme is proposed and implemented in the low-level leg control module in this paper. The proposed scheme is shown in Fig. \ref{fig_disturbance_rejection}. We establish a disturbance estimator based on the whole-body dynamics to estimate the system uncertainty, which is  compensated   by the whole-body planner.

%
%\subsection{Centroidal Dynamics}\label{se:centroidal_dynamics}
%
%
%\subsection{Whole Body Dynamics} \label{se:whole_body_dynamics}

\section{High-level Control} \label{sec:hlc}
In this section, we use the centroidal dynamics (\ref{eq:centroidal_dynamics}) for the high-level  planning via the MPC. 
%This paper presents WB-DRC that leverages whole-body dynamics to estimate and compensate for uncertainties in legged robots. However, because whole-body dynamics encompass all aspects of the robot's behavior, they introduce significant complexity. The MPC used in the high-level control module faces challenges in delivering satisfactory performance, particularly with regard to frequency and precision, with whole body dynamics. To address these limitations, we introduce the centroidal dynamics for high-level MPC planning. 
 Denote $x_c = [h\t,{L}\t, q\t]\t$, $u_c = [F_c\t, (\dot{q}^{j})\t]\t$ with $F_c = [f_1\t,\cdots,f_{n_c}\t]\t \in \mathbb{R}^{3n_c}$.  
The centroidal dynamics (\ref{eq:centroidal_dynamics}) can be rewritten as
	\begin{gather}
		\dot{x}_c = f_c(x_c,u_c) \label{eq:xc}
	\end{gather}
where $f_c(x_c,u_c)$ can be deduced from $(\ref{eq:centroidal_dynamics})$.
The linear and angular momentum are related to 
the joint velocity
via the centroidal command matrix $A(q)$ \cite{orin2013centroidal}
\begin{gather}
	\left[ {\begin{array}{*{20}{c}}
			h\\
			{L}
	\end{array}} \right] = A(q)\dot q.
\end{gather}

The continuous dynamics   (\ref{eq:xc}) is discretized over intervals of the prediction horizon $[t, t+T]$  for  the  nonlinear  trajectory optimization problem. 
 We consider the total number of steps in the MPC to be $N$, and thus the duration of each  discrete interval is $\delta t =  T/(N - 1)$. 
Define $J_c(x)$ as the terminal quadratic cost around the reference state $x_{\rm{des}}$: 
\begin{gather}
J_c(x) = (x - x_{\rm{des}}(T))\t Q_{t} (x - x_{\rm{des}}(T))\nonumber
\end{gather}
and $\varphi_c(x, u)$ as the quadratic cost around the reference state $x_{\rm{des}}$ and input $u_{\rm{des}}$:
\begin{align}
&\varphi_c(x, u) = (x - x_{\rm{des}}(T))\t  Q_{s}(x - x_{\rm{des}}(T)) \nonumber\\&+ (u - u_{\rm{des}}(T))\t R(u - u_{\rm{des}}(T)). \nonumber
\end{align}
{Let $x_{i,k}$ and $u_{i,k}$ denote the sequences of state and input variables in the MPC at the $k$-th solve, respectively, for $i = 1,\cdots,N$.} The nonlinear MPC problem can
be formulated by defining and evaluating a cost function and  constraints on the grid of nodes as 
{
\begin{align}
\text{min}_{X, U}\;\;\;\;\;\;\;&J_{c}(x_{N,k}) + \sum ^{N-1}_{k=0} \varphi_{c,i}(x_{i,k}, u_{i,k}) + l_i(x_{i,k},u_{i,k})\nonumber\\
\mbox{s.t.} \;\;\;\;\;\;\; &x_{0,k} - {x}_{m,k} = 0,\nonumber\\
\;\;\;\;\; &x_{i+1,k} - x_{i,k} - f_c(x_{i,k},u_{i,k})\delta t =0, \nonumber\\
\;\;\;\;\; &g_i(x_{i,k},u_{i,k}) = 0,\;\; i = 0,\cdots,N-1. \label{eq:MPC_problem}
\end{align}
where ${x}_{m,k}$ is the current measured state,  $X = [x_{0,k}\t, \cdots, x_{N,k}\t]\t$ and $U = [u_{0,k}\t,\cdots,u_{N-1,k}\t]\t$ are the sequences of state and input variables respectively.}  The process of establishing these constraints $l_i$ and $g_i$ follows the same approach as described in \cite{grandia2023perceptive}.  The penalized      cost function $l_i$ is derived from inequality constraints, which describe the requirement that the swing leg remains above the ground and the contact force lies within the friction cone. 
The constraint function $g_i$ captures the equality requirements that the swing leg’s end does not experience a contact force, and that the end of the stance leg has zero velocity. {Note that the legged robot is a hybrid system, where the GRFs are either zero or non-zero, as determined by the constraint $g_i$, which is managed by the gait scheduler module. }

{In centroidal dynamics (\ref{eq:centroidal_dynamics}), referred to as nominal centroidal dynamics, the force $f_i$ is considered the ground reaction force without accounting for external forces, and $m$ is treated as the robot's nominal mass, excluding model uncertainties. We use the nominal centroidal dynamics instead of its disturbed version, which incorporates external forces and model uncertainties, because we aim to generate the reference trajectory as if the system is unaffected by disturbances. }  
The low-level control will   compensate the 
the disturbance in  the whole-body dynamics (\ref{eq:dynamics}) and intends to track the   reference trajectory generated from 
the nominal  system. 

The solution to the   MPC problem (\ref{eq:MPC_problem}) is denoted as 
$X=X_{\rm{ref}} := [h_{\rm{ref}}\t, {L}_{\rm{ref}}\t, q_{\rm{ref}}\t]\t$ and  $U=U_{\rm{ref}} := [F_{\rm{ref}}\t, (\dot{q}^{j}_{\rm{ref}})\t]\t$, which gives the reference signals at the discrete time instances $[t,t+\delta_t,\cdots, t+T]\t$. 
We can   linearly interpolate the discrete  reference signals  to a continuous reference trajectory  for the interval $[t,t+T]$. The reference velocity is given by
	\begin{gather}
		\dot{q}_{\rm{ref}} = \left[ {\begin{array}{*{20}{c}}
				{\dot{q}_{\rm{ref}}^b}\\
				{\dot{q}_{\rm{ref}}^j}
		\end{array}} \right]  .
	\end{gather}  
The  reference velocity $\dot{q}^b_{\rm{ref}}$ for the base can be obtained as
\begin{gather}
\dot{q}^b_{\rm{ref}} = A^{-1}_b(q_{\rm{ref}})\left(\left[ {\begin{array}{*{20}{c}}
		h_{\rm{ref}}\\
		{L}_{\rm{ref}}
\end{array}} \right] - {A_j}({q_{\rm{ref}}}){\dot q^j_{\rm{ref}}}\right)
\end{gather}
where $A(q)$ is partitioned as $ A(q):=[A_b(q),A_j(q)]  $ with $A_b(q) \in \mathbb{R}^{6 \times 6}$. 
Partition the matrix $D(q)$ in (\ref{eq:dynamics})  as  
 \begin{gather}
 	D(q) = \left[ {\begin{array}{*{20}{c}}
 			{{D_{11}}(q)}&{{D_{12}}(q)}\\
 			{{D_{21}}(q)}&{{D_{22}}(q)}
 	\end{array}} \right]
 \end{gather}
 where $D_{11}(q) \in \mathbb{R}^{6 \times 6}$. 
The reference acceleration vector for the actuated joints ${\ddot q}^j_{\rm{ref}} \in \mathbb{R}^{n}$ is obtained by directly differentiating the   joint velocity $\dot{q}^j_{\rm{ref}}$. The  base reference acceleration $\ddot{q}^{b}_{\rm{ref}}$ for the base can be calculated as follows
\begin{align}
	\ddot{q}^b_{\rm{ref}} =& D_{11}(q_{\rm{ref}})^{-1}(S_{\theta}(-C(\dot{q}_{\rm{ref}},q_{\rm{ref}})\dot{q}_{\rm{ref}} -G(q_{\rm{ref}}) + J\t F_{\rm{ref}})\nonumber\\ &- D_{12}{(q_{ref})}{\ddot q}^j_{\rm{ref}}).\nonumber
\end{align}
where   $S_{\theta} = [I_6, 0_{6 \times n}]$.
The reference acceleration is given by
\begin{gather}
	{\ddot q_{\rm{ref}}} = \left[ {\begin{array}{*{20}{c}}
			{{{\ddot q}^b}_{\rm{ref}}}\\
			{{{\ddot q}^j}_{\rm{ref}}}
	\end{array}} \right].\label{eq:ddotqref}
\end{gather}
The input reference trajectory can be obtained by
\begin{gather}
\tau_{\rm{ref}} = S(D({q}_{\rm{ref}})\ddot{q}_{\rm{ref}} + C(q_{\rm{ref}},\dot{q}_{\rm{ref}})\dot{q}_{\rm{ref}} + G(q_{\rm{ref}}) -J\t F_{\rm{ref}})\nonumber
\end{gather}
where $S$ is given in (\ref{eq:dynamics}).

\section{Adaptive Extended State Disturbances Estimator} \label{sec:disturbance_estimator}
In this section, we first rewrite the whole-body dynamics (\ref{eq:dynamics})  and then introduce the model of uncertainties.

In (\ref{eq:dynamics}), the first six rows represent the {unactuated} dynamics, while the remaining rows correspond to the actuated dynamics.
	Let $x_1 = q$ and $x_2 = \dot{q}$. The dynamics model can be written in the state-space form 
	\begin{gather}
		\left[ {\begin{array}{*{20}{c}}
				{{{\dot x}_1}}\\
				{{{\dot x}_2}}
		\end{array}} \right] = \left[ {\begin{array}{*{20}{c}}
				{{x_2}}\\
				{{f_0}({x_1},{x_2})}
		\end{array}} \right] + \left[ {\begin{array}{*{20}{c}}
				0\\
				{{g_0}({x_1})}
		\end{array}} \right]u + \left[ {\begin{array}{*{20}{c}}
				0\\
				\bar{d}
		\end{array}} \right]
	\end{gather}
	where $\bar{d} := {{D^{ - 1}}({x_1})} d$,
	\begin{align}
		&{f_0}({x_1},{x_2}) :=  - {D^{ - 1}}({x_1})(C({x_2},{x_1})x_2 + G({x_1}))\nonumber \\
		&{g_0}({x_1}) := {D^{ - 1}}({x_1})B 
	\end{align}
	with $B = [S\t, J(q)\t]$ and $u = [\tau\t, F\t]\t$.

The uncertainties mostly come from inaccurate modeling  for   mass and   inertia, and the impact of  different terrains  on the robot is also unknown.  Inspired by \cite{sombolestan2024adaptive}, the
modeling uncertainties can be handled by
the PD control law with
	the appropriate controller parameters. As in \cite{sombolestan2024adaptive}, without directly modeling the uncertainty, 
	we propose the uncertainty in terms of 
a	PD term  with unknown parameters as follows  
\begin{gather}
	d_1(e,t) = E_q \theta \label{eq:d1}
\end{gather}
where $d_1 \in \mathbb{R}^{n}$, $\theta \in \mathbb{R}^{2n}$ are unknown parameters that need to be estimated and $E_q = \{\rm{diag}\{e_q\t\},\rm{diag}\{\dot{e}_q\t\}\} \in \mathbb{R}^{n \times 2n}$ with $e_q = q^{j}_{\rm{ref}} - q^{j}$ and $\dot{e}_q = \dot{q}^{j}_{\rm{ref}} - \dot{q}^{j}$ where both the desired joint position $q^{j}_{\rm{ref}}$ and joint velocity $\dot{q}^{j}_{\rm{ref}}$ are calculated by solving MPC problem (\ref{eq:MPC_problem}), and $q^j$ and $\dot{q}^{j}$ are joint position and velocity of the legged robot, respectively. The estimated parameter $\theta$ acts as PD controller parameters, while the uncertainty $d_1$	serves as the controller output to adjust the robot’s joint stiffness, thereby enhancing its robustness.
In addition to the  uncertainty $d_1$ in the actuated dynamics,
we introduce the additive uncertainty $d_2(t) \in \mathbb{R}^{6 + n}$  in the {unactuated} dynamics  and the uncertainty term $d$ is
\begin{gather}
	\bar{d}(e, t ) = S\t d_1(e, t) + d_2(t).  \label{eq:uncertainty_form}
\end{gather} 
As stated in \cite{zhu2023proprioceptive}, handling the uncertainty term $d_2$ would greatly improve the disturbance rejection performance  especially when the  foot contact force estimation is inaccurate.

To estimate the uncertainty $d$, recall the dynamics (\ref{eq:dynamics}) as follows:
\begin{small}
\begin{gather}
	\left[ {\begin{array}{*{20}{c}}
			{{{\dot x}_1}}\\
			{{{\dot x}_2}}
	\end{array}} \right] = \left[ {\begin{array}{*{20}{c}}
			{{x_2}}\\
			{{f_0}({x_1},{x_2})}
	\end{array}} \right] + \left[ {\begin{array}{*{20}{c}}
			0\\
			{{g_0}({x_1})}
	\end{array}} \right]u + \left[ {\begin{array}{*{20}{c}}
			0\\
			{{S\t}{E_q}\theta  + {d_2}(t)}
	\end{array}} \right]. \label{eq:dynamics_uncertainty}
\end{gather}
\end{small}
Note that there exist unknown parameters $\theta$ in $d_1$, as well as nonlinearities $d_2$, in the uncertainty $d$. We aim to propose a general whole-body disturbance rejection scheme to estimate the uncertainty in legged robot systems. To achieve this, we combine adaptive control, which is used to estimate the unknown parameters, with ESO used to estimate the nonlinearities, utilizing full-state feedback in a single unified controller. In this approach, the uncertain nonlinearities are handled simultaneously.

\subsection{ Projection Mapping and Parameter Adaptation}
\bass\label{eq:ass}
The defined parameter set $\theta$ satisfies
 
\begin{gather}
	\theta  \in {\Omega _\theta } \buildrel \Delta \over = \{ \theta = [\theta_i]:{\theta _{\rm{i}\min }} \le \theta_i  \le {\theta _{\rm{i}\max }}\} 
\end{gather}
where $\theta_{\rm{min}} = [\theta_{\rm{1min}},\cdots,\theta_{\rm{(2n)min}}]\t$ and $\theta_{\rm{max}} = [\theta_{\rm{1max}},\cdots, \theta_{\rm{(2n)max}}]\t$ are known.
\eass
Let $\hat{\theta}$ denote the estimate of $\theta$ and $\tilde{\theta}$ the estimation error (i.e., $\tilde{\theta} = {\theta} - \hat{\theta}$). A discontinuous projection can be defined as 
\begin{gather}
{\text{Proj}_{{{\hat \theta }_i}}}({ \bullet _i}) = \left\{ \begin{array}{l}
	0,\;\;\text{if}\;\;{{\hat \theta }_i} = {\theta_{i\max}}\;\;\text{and}\;\;{ \bullet _i} > 0\\
	0,\;\;\text{if}\;\;{{\hat \theta }_i} = {\theta_{i\min}}\;\;\text{and}\;\;{ \bullet _i} < 0\\
	{ \bullet _i},\;\;\text{otherwise}
\end{array} \right.
\end{gather}
where $i = 1, \cdots , 2n$. By using an adaptation law given by
\begin{gather}
\dot {\hat \theta}  = \text{Proj}{_{\hat \theta }}(\Gamma \alpha )\;\;\text{with}\;\;{\theta _{\min }} \le \hat \theta (0) \le {\theta _{\max }} \label{eq:adaptation_law}
\end{gather}
 where $\text{Proj}{_{\hat \theta }}(\Gamma \alpha ) = [\text{Proj}{_{\hat \theta_0 }}(\Gamma_0 \alpha ), \cdots, \text{Proj}{_{\hat \theta_{2n} }}(\Gamma_{2n} \alpha )]\t$, $\Gamma>0$ is a  diagonal matrix of the adaptation rate and $\Gamma_i$ represents the $i$-th row of the matrix $\Gamma$,  and $\alpha$ is an adaptation function to be synthesized later; for any adaptation function $\alpha$, the projection mapping guarantees \cite{yao1996smooth}
\begin{gather}
	\hat \theta  \in {\Omega _{\hat \theta }} \buildrel \Delta \over = \left\{ {\hat \theta :{\theta _{\min }} \le \hat \theta  \le {\theta _{\max }}} \right\}\nonumber\\
	{ {{\tilde \theta }\t}[{\Gamma ^{ - 1}}\rm{Proj}_{\hat \theta }(\Gamma \alpha ) + \alpha ] \le 0,\forall \alpha}. \label{eq:thetaBound}
\end{gather}
%{\blue In (\ref{eq:thetaBound}), $\theta_{\rm{imin}}$ should be set to a value less than zero, and $\theta_{\rm{imax}}$ should be set to a value greater than zero for $i = 1,\cdots 2n$. }
The adaptation rate $\Gamma$ controls the balance between responsiveness and stability, with a higher rate enabling quicker adjustments at the cost of stability, and a lower rate ensuring stability but slower responsiveness.
\subsection{ESO Design}\label{sec:eso}
We introduce the extended state $x_3 \in \mathbb{R}^{6 + n}$ as $x_3 = d_2$ and $h$ denote the rate of change of $x_3$, i.e., $h= \dot{x}_3$. Both $x_3$ and $h$ are assumed to be unknown but bounded  signal vectors. The system (\ref{eq:dynamics_uncertainty}) can be expressed in a state-space form as 
\begin{align}
\left\{ {\begin{array}{*{20}{l}}
			{{\dot x}_1} = {x_2}\\
			{{\dot x}_2} = {f_0}({x_1},{x_2}) + {g_0}({x_1})u + {S\t}{E_q}\theta  + {x_3}\\
		{{{\dot x}_3} = h(t)}
\end{array}} \right. \label{eq:state_equation_form}
\end{align}

\brem
There exists another way to define the extend state $x_3$, as $x_3 = S\t E_q\tilde{\theta} + d_2(t)$. Then, the system (\ref{eq:dynamics_uncertainty}) can be rewritten as
\begin{gather}
\left\{ {\begin{array}{*{20}{l}}
			{{\dot x}_1} = {x_2}\\
			{{\dot x}_2} = {f_0}({x_1},{x_2}) + {g_0}({x_1})u + {S\t}{E_q}\hat \theta  + {x_3}\\
		{{{\dot x}_3} = h(t)}\nonumber
\end{array}} \right.
\end{gather}
It is worth noting that,  the two definitions of the extended states do not affect the  ESO to be constructed later.  Although   the dynamics of the state estimation errors will be different, the subsequent stability analysis and conclusions remain the same.
\erem

Let $\hat{x}_1$, $\hat{x}_2$, and $\hat{x}_3$ be the estimate of $x_1$, $x_2$ and $x_3$, respectively.
By referring to \cite{1242516}, the linear ESO can be constructed as follows:
\begin{align}
\left\{ {\begin{array}{*{20}{l}}
			{{\dot {\hat x}}_1} = {{\hat x}_2} + 3{\omega _0}({x_1} - {{\hat x}_1})\\
			{{\dot {\hat x}}_2} = {f_0}({x_1},{x_2}) + {g_0}({x_1})\hat u + {S\t}{E_q}\hat \theta  + {{\hat x}_3} + 3{\omega^2 _0}({x_1} - {{\hat x}_1})\\
		{{{\dot {\hat x}}_3} = {\omega^3 _0}({x_1} - {{\hat x}_1})}
\end{array}} \right.\label{eq:ESO}
\end{align}
where $\hat{u} = [F_{\rm{ref}}\t, \tau\t]\t$ with $F_{\rm{ref}}$ being obtained by solving the MPC problem (\ref{eq:MPC_problem}), and $\omega_0$ are tuning parameters to be determined.   The parameter  $\omega_0$ affects the bandwidths of the  ESO.  While a large $\omega_0$ is necessary for fast disturbance estimation, it can also amplify noise.
Due to the absence of ground reaction force sensors on the legged robot, we use $F_{\rm{ref}}$ in $\hat{u}$ to to replace the actual GRFs $F$ in $u$, and the discrepancy between $F$ and $F_{\rm{ref}}$ is estimated by ESO.
Note that if the legged robot operates without uncertainty, and the robot's position $q$ and velocity $\dot{q}$ perfectly track ${q}_{\rm{ref}}$ and $\dot{q}_{\rm{ref}}$, then $\hat{u}$ will be equal to $u$, i.e., $u - \hat{u} = {0}$. For the ESO in (\ref{eq:ESO}), the system state $x$ and system input $\tau$ in $\hat{u}$ can be directly measured or obtained by state estimation, while $\hat{\theta}$ is computed using the adaptation law (\ref{eq:adaptation_law}).

Next, we will analyze the stability of the proposed ESO. Let 
\begin{equation}
	\eta = [(x_1 - \hat{x}_1)\t, (x_2 - \hat{x}_2)\t/\omega_0, (x_3 - \hat{x}_3)\t/\omega^2_0]\t \label{eq:eta}
\end{equation}
 be the state estimation error and $ \tilde{u} = u - \hat{u}$. The dynamics of the state estimation errors can be expressed as 
\begin{gather}
\dot \eta  = {\omega _0}A\eta  + \frac{{{C_1}({g_0}({x_1})\tilde u + {S\t}{E_q}\tilde \theta )}}{{{\omega _0}}} + \frac{{{C_2}h(t)}}{\omega^2 _0}\label{eq:eta_dynamics}
\end{gather}
where $\tilde{u} = u - \hat{u}$ is an unknown but bounded vector, since $||\tilde{u}|| = ||F- F_{\rm{ref}}||$, with $F_{\rm{ref}}$ being bounded as a result of solving MPC problem (\ref{eq:MPC_problem}), $g_0(x_1)$ is a bounded matrix, and
\begin{gather}
A = \left[ {\begin{array}{*{20}{c}}
		{ - 3}&1&0\\
		{ - 3}&0&1\\
		{ - 1}&0&0
\end{array}} \right] \otimes I_{6 + n},\;\;{C_1} = \left[ {\begin{array}{*{20}{c}}
0\\
1\\
0
\end{array}} \right]\otimes I_{6 + n},\nonumber\\{C_2} = \left[ {\begin{array}{*{20}{c}}
0\\
0\\
1
\end{array}} \right]\otimes I_{6 + n}.
\end{gather}

\bthm\label{thm:them1} Consider the error dynamics (\ref{eq:eta_dynamics}) and   adaptation dynamics (\ref{eq:adaptation_law})  under Assumption \ref{eq:ass} where  the  adaptation function $\alpha $
is
 \begin{gather}
	\alpha = \frac{E_q\t S C_1\t P \eta}{\omega_0} \label{eq:adaptation_law2}
\end{gather}
with ESO bandwidth parameter $\omega_0 > 0$ and matrix $P$ satisfying $A\t P + PA = -I$.
Suppose there is no   uncertainty, i.e., $x_3 = 0$ and $h = 0$. When the robot's state $x = \text{col}(q,\dot{q})$   perfectly tracks the reference trajectory $x_{\rm{ref}} = \text{col}(q_{\rm{ref}}, \dot{q}_{\rm{ref}})$ derived from the MPC problem (\ref{eq:MPC_problem}), i.e., $\tilde{u} = {0}$,  one has
$\lim_{t \to \infty}( x - \hat{x}) = {0}$.   
% 
%
%such the error system ($\ref{eq:eta_dynamics}$) is asymptotically stable, i.e., $\lim_{t \to \infty}( x - \hat{x}) = {0}$.   
\ethm

\proofnow
Let us consider the following control Lyapunov candidate function:
\begin{gather}
V(\eta ,\tilde \theta ) = {\eta \t}P\eta  + \tilde{\theta}\t {\Gamma ^{ - 1}}{\tilde{\theta}} .\label{eq:V}
\end{gather}
Therefore, its time derivative will be
\begin{align}
	\dot V(\eta ,\tilde \theta ) =&  - {\omega _0}||\eta |{|^2} + {{\tilde \theta }\t}({\Gamma ^{ - 1}}\dot {\tilde \theta}  + \frac{ {E_q\t} SC_1\t P\eta}{\omega_0} ) \nonumber \\+& ({{\dot {\tilde \theta} }\t}{\Gamma ^{ - 1}} +\frac{{\eta \t}PC_1{S\t}{E_q}}{\omega_0})\tilde \theta +
	\frac{{{\tilde u}\t}{g_0\t}{C_1\t} P\eta  + {\eta\t}P{C_1}{g_0}\tilde u}{\omega_0} \nonumber\\+ &\frac{{{h\t}{C_2\t}P\eta  + {\eta\t}P{C_2}h}}{{{\omega^2 _0}}}.\label{eq:dotV}
\end{align}
Combining the adaptation law (\ref{eq:adaptation_law2}) and (\ref{eq:dotV}), we have 
\begin{gather}
\dot{V}(\eta,\tilde{\theta}) = -\omega_0||\eta||^2.\label{eq:dotV2}
\end{gather}
Combining (\ref{eq:V}) and (\ref{eq:dotV2}), it gives $\eta$ and $\tilde{\theta}$ are bounded. According to (\ref{eq:eta_dynamics}), we have $\dot{\eta}$ is bounded. Using Barbalat Lemma, (\ref{eq:dotV2}) implies $\lim_{t \to \infty}( x - \hat{x}) = {0}$. 
\eproof

%{ \red This remark is very confusing.}
%\brem
%In the adaptation law given by (\ref{eq:adaptation_law2}), the estimation error vector $\eta$ is related to the $x_3$. Specifically, the quantity $x_3 = 0$ is known under the condition of Theorem \ref{thm:them1}. Based on this, the adaptation law (\ref{eq:adaptation_law2}) can be determined accordingly.
%\erem

\bthm
Consider the error dynamics (\ref{eq:eta_dynamics}) and   adaptation dynamics (\ref{eq:adaptation_law})  under Assumption \ref{eq:ass} where the  adaptation function $\alpha $
is
\begin{gather}
	\alpha = \frac{E_q\t S C_1\t P \bar{\eta}}{\omega_0}\nonumber
\end{gather}
where $\omega_0 > 0$, $\bar{\eta} = [(x_1 - \hat{x}_1)\t, (x_2 - \hat{x}_2)\t/\omega_0, - \hat{x}_3\t/\omega^2_0]\t$, and the matrix $P$ satisfying $A \t P + PA = -I$, $||\tilde{\theta}|| < \tilde{\theta}_b$.
Suppose  both the disturbance $x_3$ and its derivative $h$ are bounded, i.e., $||x_3|| \le d_b$   and $||h|| < h_b$.
If
 both the matrix $E_q$ and the  error vector $g_0\tilde{u}$ are bounded, i.e., $||E_q|| < e_b$, $||g_0\tilde{u}|| \le \tilde{u}_b$, then the estimation error $\eta$ is bounded.
\ethm
\proofnow
By Assumption \ref{eq:ass} and property of the projection operators, the projection operators will keep $\tilde{\theta}$ bounded. We have these bounds as
$
||\tilde{\theta}|| < \tilde{\theta}_b.
$
From (\ref{eq:V}), we have 
\begin{gather}
	V(\eta ,\tilde \theta) \le {\lambda _{\max }}(P)||\eta||^2 + {||\Gamma |{|^{ - 1}}}{\tilde{\theta}_b}^2
	\label{eq:V_1}
\end{gather} with 
	$\lambda_{\max }(P)$ being the maximum eigenvalue of matrix $P$. From (\ref{eq:dotV}), we have
{
\begin{align}
\dot V(\eta ,\tilde \theta ) \le&  - {\omega _0}||\eta |{|^2} + { \left(\frac{2{\tilde u_b}||{C_1\t}P||}{\omega_0} + \frac{{2h_b||{C_2\t}P||}}{{{\omega^2 _0}}}\right)}||\eta ||\nonumber\\
&+ \frac{2 ||SC_1\t P||e_bd_b\tilde{\theta}_b}{\omega^3_0}{ .}
\label{eq:dotV1}
\end{align}
}
Let
$
\lambda = {\omega_0}/({2\lambda_{\max }(P)}).
$ Combining (\ref{eq:V_1}) and (\ref{eq:dotV1}), we have
\begin{gather}
	\dot{V}(\eta ,\tilde \theta ) + \lambda  V(\eta ,\tilde \theta ) \le a{(||\eta || + \frac{b}{{2}})^2} + \lambda {\delta _V}\label{eq:V_add_dotV}
\end{gather}
where 
\begin{align}
a =& -\omega_0 + \lambda \lambda_{\rm{max}}(P),\;\; b = 2{ \left(\frac{{\tilde u_b}||{C_1\t}P||}{\omega_0} + \frac{{h_b||{C_2\t}P||}}{{{\omega^2 _0}}}\right)}/a,\nonumber\\
	{\delta _V} =& \frac{{2{{({\omega _0}{{\tilde u}_b}||{C_1\t}P|| + {h_b}||{C_2\t}P||)}^2}}}{{\lambda{\omega _0}^5}}  + \frac{2 ||SC_1\t P||e_bd_b\tilde{\theta}_b}{\lambda\omega^3_0} \nonumber\\ &+ ||\Gamma |{|^{ - 1}}{\tilde \theta _b}.\nonumber
\end{align}
From (\ref{eq:V_add_dotV}), we have
\begin{align}
	\dot V(\eta ,\tilde \theta ) + \lambda V(\eta ,\tilde \theta ) \le \lambda {\delta _V}.\label{eq:V_add_lamdaV}
\end{align}
It follows that 
\begin{gather}
V(\eta,\tilde{\theta}) \le V(\eta_0,\tilde{\theta}_0)\exp ( - \lambda t) + {{\delta _V}}(1 - \exp ( - \lambda t))\nonumber
\end{gather}
where $\eta_0$ and $\tilde{\theta}_0$ are the  values of $\eta$ and $\tilde{\theta}$, respectively.

Since $V(\eta,\tilde{\theta}) \ge \lambda_{\min }(P)||\eta||^2$, with $\lambda_{\min }(P)$ being the minimum eigenvalue of matrix $P$, it follows that
\begin{gather}
||\eta || \le \sqrt {\frac{{V({\eta _0},{{\tilde \theta }_0})\exp ( - \lambda t) + {\delta _V}(1 - \exp ( - \lambda t))}}{{{\lambda _{\min }}(P)}}} . \nonumber
\end{gather}
Thus, the estimation error $\eta$ is bounded. 
%Furthermore,  by choosing the adaptation gain $\Gamma$ and ESO bandwidths $\omega_0$ sufficiently large, we can limit  $\eta$ in an arbitrarily small neighborhood $\delta_V$ of the origin.
\eproof

%{\red why boundedness is so important that you need to present these two lemmas?} 

%\subsection{Kalman Filter Design}
%
%Rewrite the ESO as 
%\begin{gather}
%\left\{ \begin{array}{l}
%	\dot {\hat x} = A\hat x + w(q, \dot{q}, \hat{\theta}, \hat{u})\\
%	{y_m} = {C_m}\hat x + v
%\end{array} \right. \label{eq:hatx_dynamics}
%\end{gather}
%where
%\begin{align}
%&d(\hat \theta ,\varphi ,x) = \left[ {\begin{array}{*{20}{c}}
%		{3{\omega _0}{x_1}}\\
%		{{f_2}(x) + \hat \theta\t {E_q} + 3{\omega _0}^2{x_1} + {g_2}(x)}\\
%		{\omega _0^3{x_1}}
%\end{array}} \right]\nonumber\\&{C_m} = \left[ {\begin{array}{*{20}{c}}
%0&0&1
%\end{array}} \right]
%\end{align}
%We design the following Kalman filter:
%\begin{gather}
%\dot z = Az + {K_s}({y_m} - {C_m}z) \label{eq:z_dynamics}
%\end{gather}
%Let the state estimation error of the KF be $\chi = \hat{x} - z$, then the derivative of $\chi$ is obtained from (\ref{eq:hatx_dynamics})
%and (\ref{eq:z_dynamics}) as 
%\begin{gather}
%\dot{\chi} = A_k\chi + d(\hat{\theta},\varphi,x)
%\end{gather}
%where $A_k = A - K_sC_m$.
%
%\bthm
%The state estimates, obtained by the KF, are optimal if the KF gain vector is
%\begin{align}
%K =& PC_m\t R^{-1}\nonumber\\
%\dot{P} =& - PC_m\t R^{-1} C_mP +AP + PA\t + B_dQ_dB_d\t
%\end{align}
%where $P$ is the state estimation-error covariance matrix.
%\ethm

\section{Disturbance Rejection Whole Body Planner}\label{sec:disturbance_rejction_WBC}
The uncertainty of the robot system is estimated in Section \ref{sec:eso}. In this section, we will focus on compensating for the uncertainty in the robot system by using a quadratic programming and WBC framework (QP-WBC).  

%First, we will compute the desired GRFs by solving a quadratic programming (QP) optimization problem. Second, the WBC framework will be used to track both the GRFs obtained from the QP problem and the reference acceleration in (\ref{eq:ddotqref}).   

To reduce the noise, we applied a moving average filter (MAF) to smooth the estimated uncertainty. The estimated uncertainty given by the adaptive ESO is
\begin{gather}
	\hat{f} = D(x_1)(\hat{x}_3 + S\t E_q \hat{\theta}) . \label{eq:estimated_uncertainty}
\end{gather}  
Let  $\hat{f}_{\rm{filter}}$ be the moving average filtered value of $\hat f$ at this moment.

Next, we use the hierarchical WBC to address uncertainties in the legged system. A detailed formulation of the WBC can be found in \cite{bellicoso2016perception}. Each task within the WBC is represented as an equality constraint or an inequality constraint on generalized accelerations, joint actuation torques, and GRFs. Define the desired wrench of the robot as 
	\begin{gather}
		W^{*}_d =  \hat{f}_{\rm{filter}} + J(q_{\rm{ref}})\t F_{\rm{ref}} + S\t \tau_{\rm{ref}}\label{eq:wd}
	\end{gather}
which is the value of   the right-hand side of the equation in dynamics (\ref{eq:dynamics})  when the robot's position, velocity, and acceleration perfectly track the corresponding reference values and the disturbance is well estimated.
To formulate the WBC problem, the dynamic constraint in WBC is expressed as:
\begin{gather}
D(q)\ddot{q}_d + C(q,\dot{q})\dot{q} + G(q) = W^*_d
\end{gather}
where $\ddot{q}_d$ represents the optimal acceleration required in the WBC, and $q$ and $\dot{q}$ are the measured joint position and velocity.
% Note that the uncertainty and the desired GRFs are accounted for in the desired wrench $W^*_d$.  
 The GRFs of the legged robot vary with changes in uncertainty, meaning that we cannot directly use the reference GRFs $F_{\rm{ref}}$ obtained from the MPC problem based on the ideal dynamics, to optimize the GRFs in the WBC. Inspired by the method in \cite{zhu2023proprioceptive} for calculating GRFs with uncertainty in legged robots, we formulate a QP problem to obtain the desired GRFs $F^*_r$:
\begin{align}
{\rm{min}}_{\tau_r, F_r}&\frac{1}{2}||F_r - F_{\rm{ref}}||^2_{Q_1} + \frac{1}{2}||J(q_{\rm{ref}})\t F_r + S\t\tau_r - W^{*}_d||^2_{Q_2}\nonumber\\
&\text{s.t.} \quad \quad C_fF_r \le 0  \label{eq:qp}
\end{align}
where $C_f$ is the contact constraint matrix, and $Q_1$ and $Q_2$ are the
weight matrices for these two objectives respectively.  In this QP problem (\ref{eq:qp}), we obtain the desired GRFs $F^*_r$ to ensure that the wrench of the legged robot, without uncertainty (i.e., the right-hand side of the dynamics equation (\ref{eq:dynamics}) with $d = 0$), matches the desired wrench $W^*_d$, with $F^*_r$ approximating the reference GRFs $F_{\rm{ref}}$. Let $\hat{f}^{w}_{\rm{filter}} = \hat{f}_{\rm{filter}}-J(q_{\rm{ref}})\t(F^*_{r} - F_{\rm{ref}})$, where $F^*_r$ corresponds to the solution of the QP.

Then, the desired wrench (\ref{eq:wd}) can be rewritten as
 	\begin{gather}
 		W^*_d = J(q_{\rm{ref}})\t F^*_{r} + S\t \tau_{\rm{ref}} + \hat{f}^{w}_{\rm{filter}} . \label{eq:wd1}
 	\end{gather}
  Note that the wrench in (\ref{eq:wd}) is equivalent to that in (\ref{eq:wd1}). The difference lies in the introduction of the GRFs $F^{*}_{r}$, given by (\ref{eq:wd1}),  which are used for tracking in the WBC, a feature not present in (\ref{eq:wd}). These desired GRFs $F^*_{r}$ will later serve as the foot-end force tracking constraint in the designed WBC, while  $\hat{f}^{w}_{\rm{filter}}$ obtained by (\ref{eq:wd1}) will serve as dynamics constraint.

The WBC can be formulated as 
\begin{align}
	\text{find}\;\;&\ddot{q}_d, F_d, \tau_d\nonumber\\
	\text{s.t.}\;\;\;\; &\underline{u} \le u \le \overline{u} \quad \quad \quad \quad \quad \quad \;\;\quad \quad \quad\quad\quad \quad \text{(Torque limits)}\nonumber\\
	&D(q)\ddot{q}_d + C(q,\dot{q})\dot{q} + G(q) = S\t \tau_d + J(q)\t F_d + \hat{f}^{w}_{\rm{filter}}  \;\nonumber\\\nonumber &\quad \quad \quad \quad  \quad \quad \quad \quad \quad \quad \quad \quad 
	\quad \quad \quad \quad \quad \quad \text{(Dynamics)}\nonumber\\
	&J(q)\ddot{q}_d + \dot{J}(q)\dot{q} = 0  \quad \quad  \text{(No motion for contact foots)}\nonumber\\
	&C_fF_d \le 0 \quad \quad \quad \quad \quad \quad \quad \quad \quad \quad \quad \quad \;\; \text{(Friction cone)} \nonumber\\
	&S_{w}F_d = 0\quad \quad \;\quad \quad \quad  \quad \; \text{(No wrench for swing foots)}\nonumber\\
	&\ddot{q}_d - \ddot{q}_{\rm{ref}} = 0 \quad \quad \quad \quad \quad \quad \quad \;\;\; \text{(Acceleration tracking)} \nonumber\\
	&F_d - F^*_r = 0 \;\; \quad \quad \quad \quad \quad \quad \;\;  \text{(Foot-end force tracking)}\nonumber
\end{align}
where $\underline{u}$ and $\overline{u}$ represent the lower and upper limits of the torque, $q$ and $\dot{q}$ are measured joint position and velocity,  $S_w$ is the swing leg selection matrix ensuring that the contact forces of the swing leg are selected. $\ddot{q}_{\rm{ref}}$ represents the desired acceleration and is calculated in (\ref{eq:ddotqref}). Tasks are solved in a strict prioritized order. The priority of  tasks is described in Table \ref{table:task}. Finally, the actuation joint torque $u_d$ can be computed using the PD control law as follows: 
\begin{gather}
	u_d = \tau^*_d + K_p(q^j_{\rm{ref}} - q^j) + K_d(\dot{q}^j_{\rm{ref}} - \dot{q}^j)
\end{gather}
where $\tau^*_d$ is obtained by solving WBC problem, $K_p \in \mathbb{R}^{n \times n}$ and $K_d \in \mathbb{R}^{n \times n}$ are positive-define diagonal matrices of
proportional and derivative gains, respectively, 
 $q^j_{\rm{ref}}$ and $\dot{q}^j_{\rm{ref}}$ are the bottom $n$ rows of $q_{\rm{ref}}$ and $\dot{q}_{\rm{ref}}$, respectively.

\begin{table}[h]
\begin{center}
\caption{Priority of Tasks}
\begin{tabular}{l|l}
\hline
Priority & Task\\
\hline
1 & Dynamics + Friction cone + No wrench for swings\\&+ Torque limits	\\
\hline
2 & Acceleration tracking\\
\hline
3 & No motion for contact foots + Foot-end force tracking\\
\hline
\end{tabular}\label{table:task}
\end{center}
\end{table}

\section{Simulation Comparison} \label{sec:simulation}
To assess the effectiveness and versatility of the WB-DRC, we conduct a series of simulation experiments on legged robots and compare the performance of the proposed WB-DRC with the widely adopted standard WBC from \cite{grandia2023perceptive}.  All simulations are performed on a single PC (Intel i7-13700KF, 3.4 GHz). For the simulations, the control system is implemented using ROS Noetic in the Gazebo simulator. The simulations are carried out on both a   biped robot, which has $n = 12$ degrees of actuation freedom, and the Unitree A1 quadruped robot, which also has $n = 12$ degrees of actuation freedom. The biped robot simulation is detailed in Section \ref{sec:simulation_biped_robot}, while the quadruped robot simulation is described in Section \ref{sec:simulation_quadruped_robot}.

\subsection{Simulation on Biped Robot} \label{sec:simulation_biped_robot}

\begin{figure}[!ht]
	\centering
	\begin{minipage}{0.38\linewidth}
		\centering
		\includegraphics[width=0.9\linewidth]{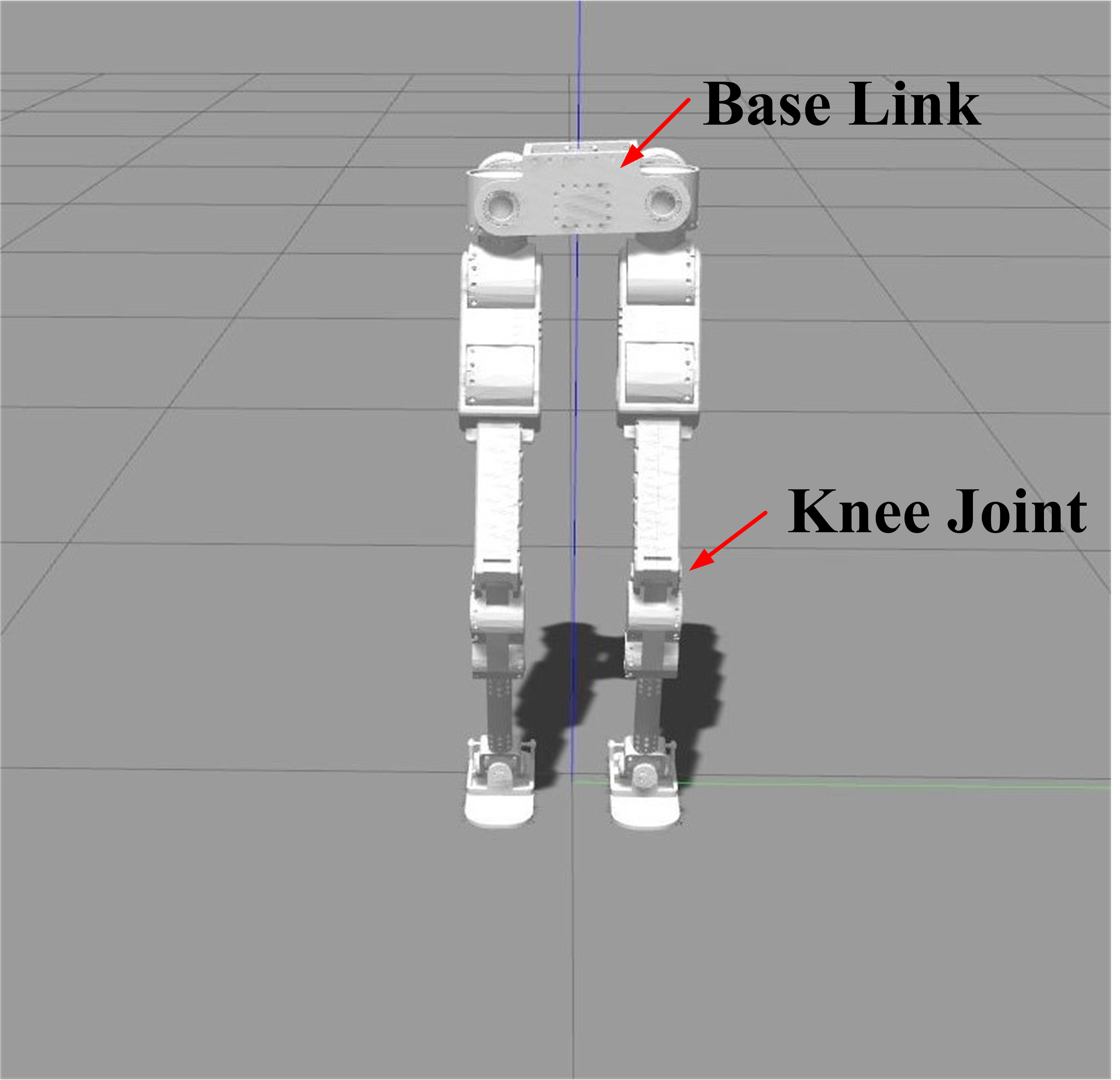}
	\end{minipage}
	\begin{minipage}{0.38\linewidth}
		\centering
		\includegraphics[width=0.9\linewidth]{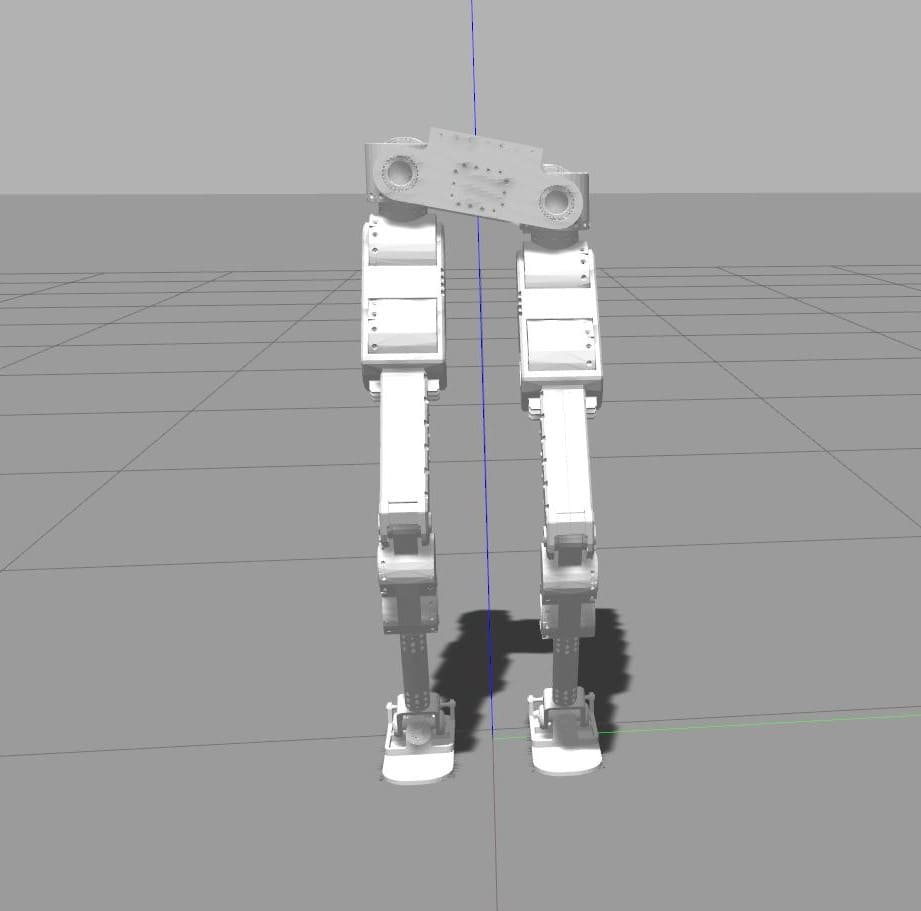}
	\end{minipage}
	\caption{Biped robot with WB-DRC (left) and with standard WBC (right) in a simulation environment.}
	\label{fig_biped_gazebo}
\end{figure}

\begin{figure}[!ht]
	\centering
	\includegraphics[scale=0.027]{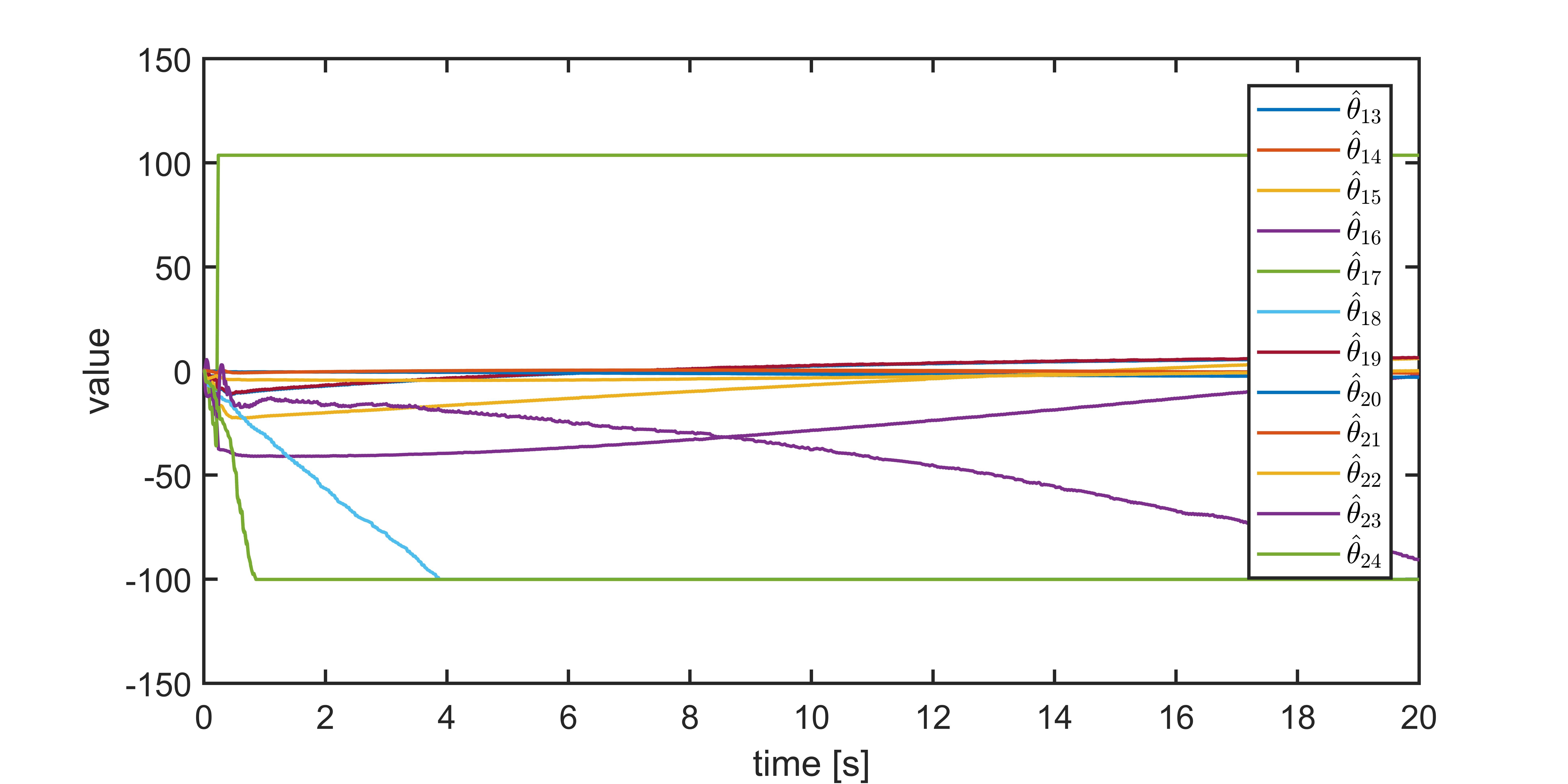}
	\caption{The estimation value $\hat{\theta}_{13} \sim\hat{\theta}_{24}$  using adaptive law  when the output torque of the robot knee joint is reduced by $30$\%.}
	\label{fig_hatTheta}
\end{figure}

The operating frequency of the MPC is $ 50 $Hz, with a prediction horizon $T = 1$ s and the total number of steps $N = 30$, while the operating frequency of   the adaptive ESO and QP-WBC are $1000 $Hz. The sliding window width of the MAF is set to $7$. The control parameters for the WB-DRC are specified in Table \ref{table:parameter_biped}. The desired height of base link (see Fig. \ref{fig_biped_gazebo}) is set to $0.86$ m. To verify the robustness of the proposed WB-DRC and the effectiveness of the disturbance estimator, we simulated a scenario in Gazebo by deliberately reducing   the knee joint torque by  $30$\%. The   standing pose with the WB-DRC and that with the standard WBC in simulation environment are shown in Fig. \ref{fig_biped_gazebo}.
It demonstrates that
the biped robot is capable of handling output torque  reduction through WB-DRC.  The estimation parameters $\hat{\theta}$ for the robot in WB-DRC using the adaptation law are presented in Fig. \ref{fig_hatTheta}. All estimated values, $\hat{\theta}_{13}\sim\hat{\theta}_{24}$,  remain bounded within the interval $[-100,100]$. 

\begin{figure}[!ht]
	\centering
	\includegraphics[scale=0.027]{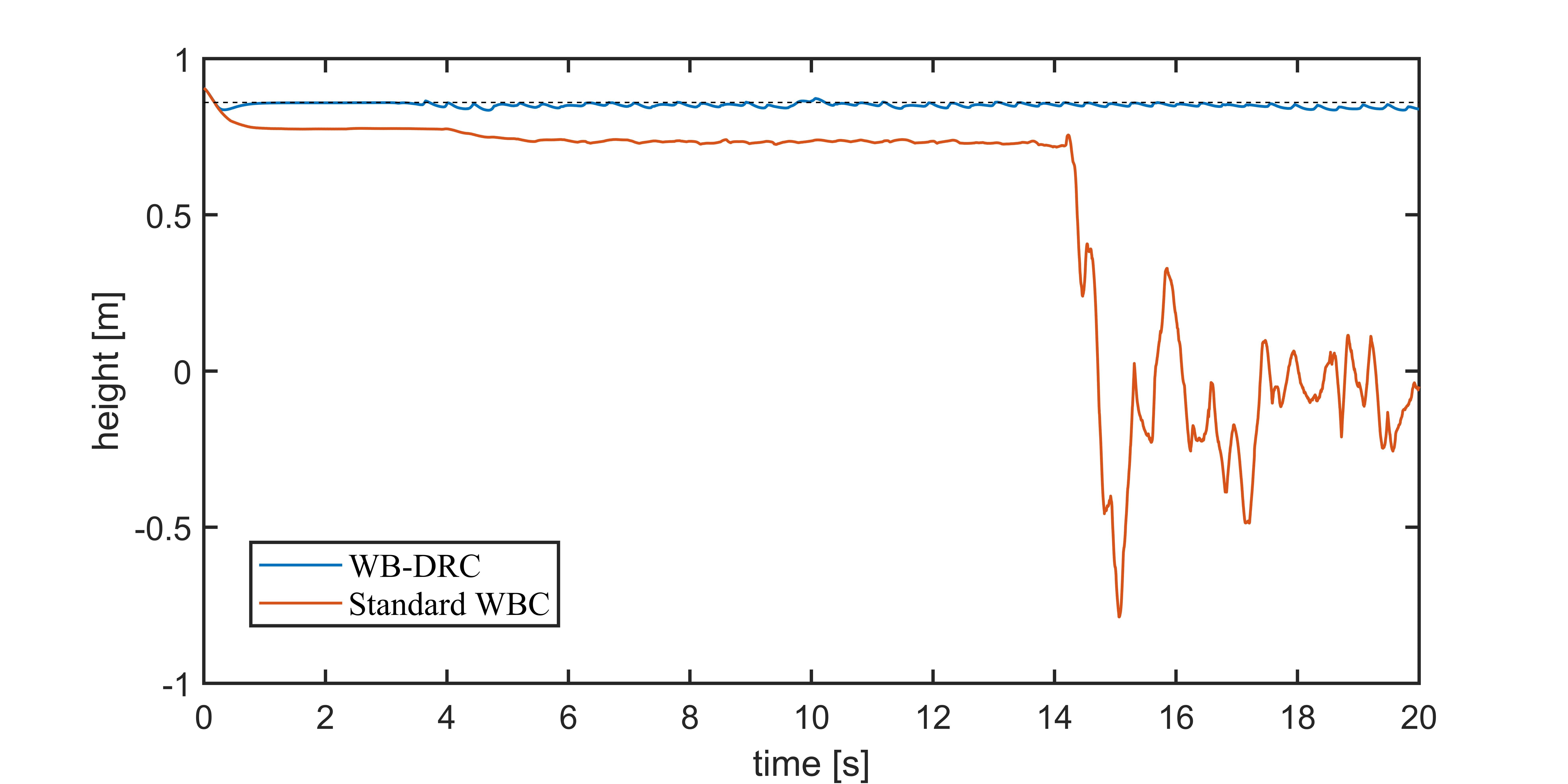}
	\caption{The height of the base link from the ground over time as the robot performs stepping in place with an 8 kg load.}
	\label{fig_model_uncertainty_biped}
\end{figure}

\begin{figure}[!ht]
	\centering
	\includegraphics[scale=0.027]{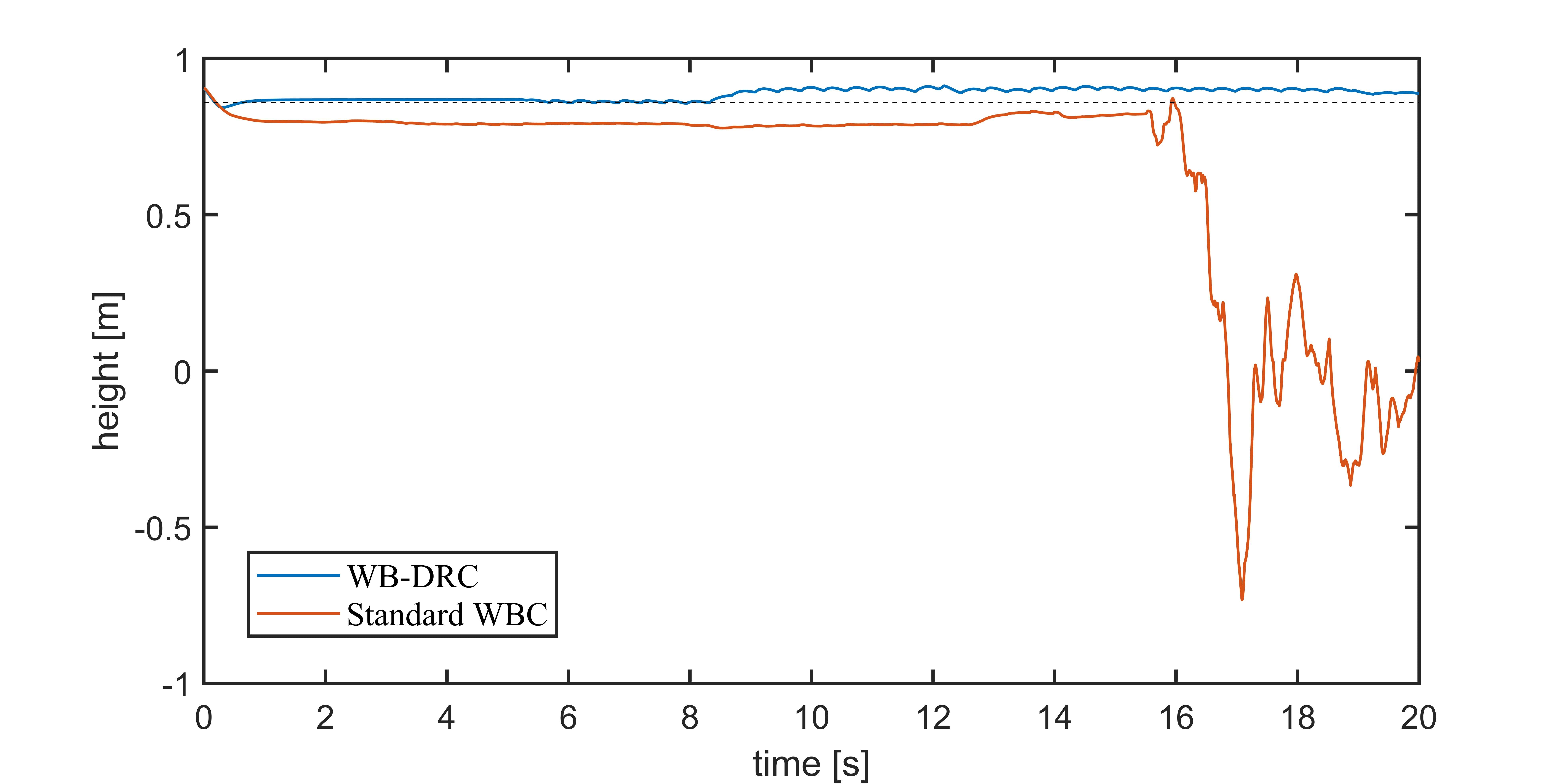}
	\caption{The height of the base link from the ground over time as the robot walks with an 6 kg load.}
	\label{fig_model_uncertainty_biped1}
\end{figure}

Fig. \ref{fig_model_uncertainty_biped} illustrates the height of the base link from the ground over time as the robot performs the stepping in place with an $8$ kg load, from $4$ s to $20$ s. The results show that the base link height of the biped robot using WB-DRC closely tracks the desired height. However, the robot with the standard WBC does not tracks the desired height and falls to the ground at around $14$s. Similarly, Fig. \ref{fig_model_uncertainty_biped1} shows the base link height as the robot walks with a $6$ kg load at $8$ s. It shows that   the robot using the standard WBC falls at around $16$ s. These results suggest that the biped robot with WB-DRC is more robust to the  model uncertainty than the robot with the standard WBC.

\begin{table}[h]
	\begin{center}
		\caption{WB-DRC Setting}
		\begin{tabular}{l|l|l}
			\hline
			Parameter & Value for Biped Robot & Value for Quadruped Robot\\
			\hline
			$\Gamma$ & $50000\times I_{ 24 \times 24 }$& $6 \times 10^5 \times I_{ 24 \times 24 }$\\
			\hline
			$\omega_0$ & $100$ & $350$ \\
			\hline
			$\hat{\theta}(0)$ & $ \textbf{0}_{24 \times 1}$& $ \textbf{0}_{24 \times 1}$\\
			\hline
			$\theta_{\rm{max}}$ & $100 \times \textbf{1}_{24 \times 1}$ & $ 100 \times \textbf{1}_{24 \times 1}$ \\
			\hline
			$\theta_{\rm{min}}$ & $-100 \times \textbf{1}_{24 \times 1}$ &  $-100 \times \textbf{1}_{24 \times 1}$\\
			\hline
			$Q_1$ & $100 \times I_{ 24 \times 24 }$ &  $100 \times I_{ 24 \times 24 }$\\
			\hline
			$Q_2$ & $I_{ 18 \times 18 }$ & $I_{ 18 \times 18 }$ \\
			\hline
		{$K_p$} & {$15 \times I_{12 \times 12}$} & {$0 \times I_{12 \times 12}$}\\
			\hline
			$K_d$ & { $0.5 \times I_{12 \times 12}$} & {$3 \times I_{12 \times 12}$} \\
			\hline
		\end{tabular}\label{table:parameter_biped}
	\end{center}
\end{table}

\subsection{Simulation on Quadruped Robot}\label{sec:simulation_quadruped_robot}
\begin{figure}[!ht]
	\centering
	\begin{minipage}{0.26\linewidth}
		\centering
		\includegraphics[width=0.9\linewidth]{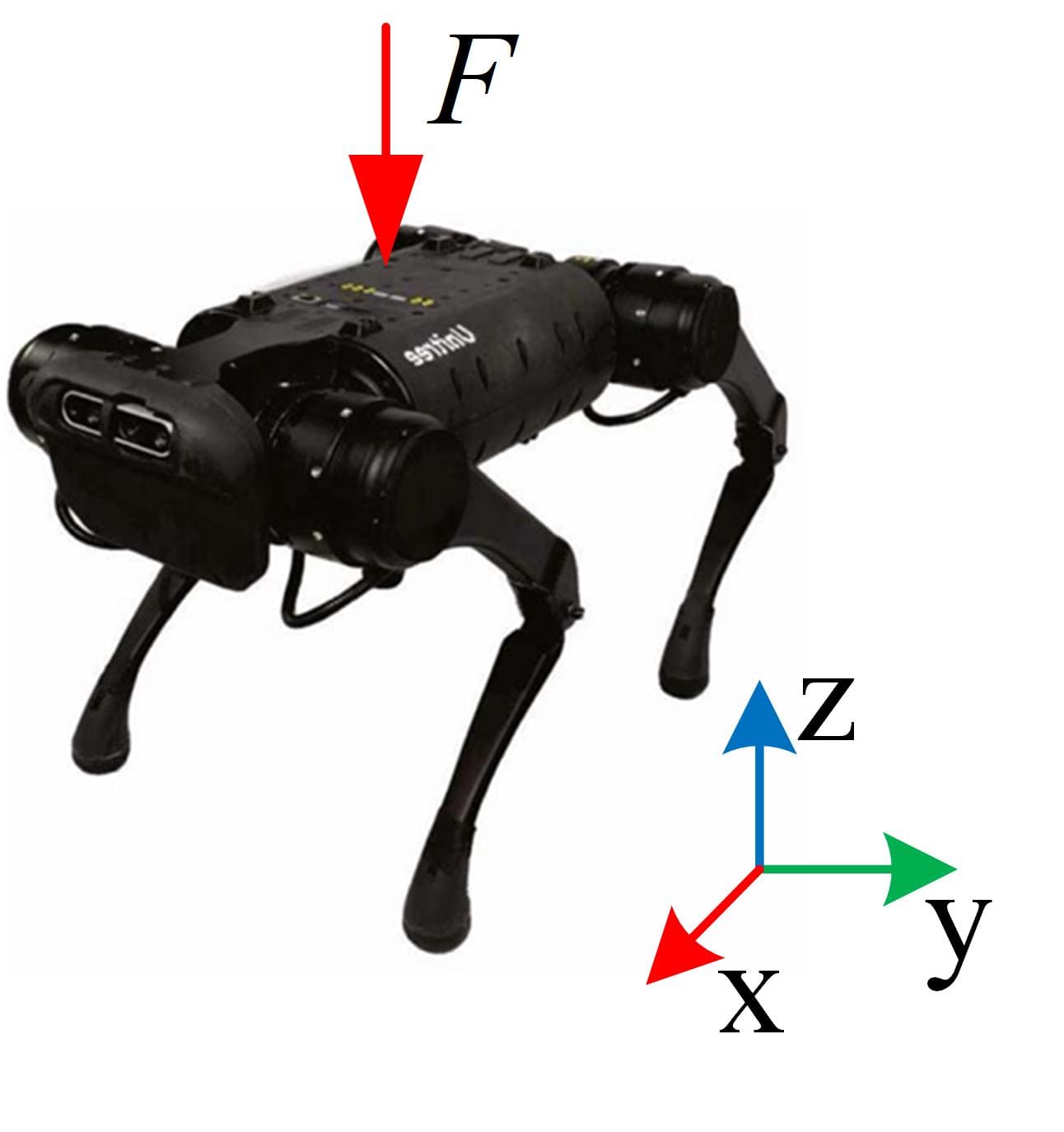}
		\label{chutian1}
	\end{minipage}
	%\qquad
	\begin{minipage}{0.56\linewidth}
		\centering
		\includegraphics[width=0.9\linewidth]{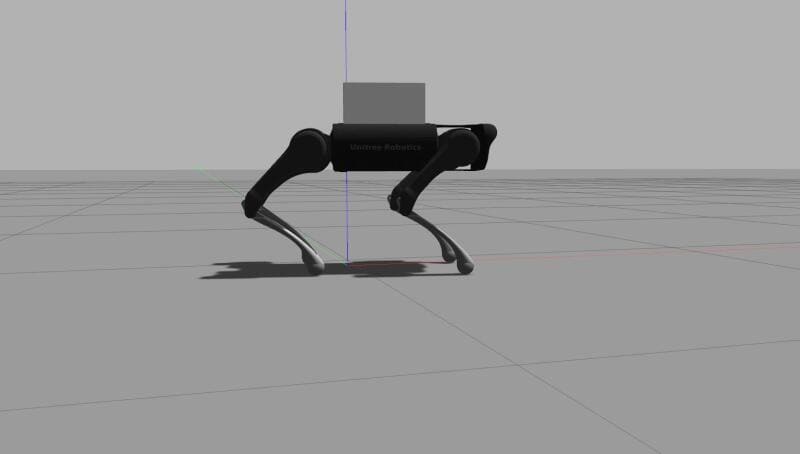}
		\label{chutian2}
	\end{minipage}
	\caption{The Unitree A1 robot (left) and its model in the simulation environment with an $8$ kg load (right).}
	\label{A1_gazebo}
\end{figure}

The operating frequency of the MPC is $20 $Hz {with a prediction horizon $T = 1$ s and the total number of steps $N = 30$,} and the running frequency of the adaptive ESO and QP-WBC are $1000$ Hz. The sliding window width of MAF is set to $3$. We set the control parameters for WB-DRC as presented in Table \ref{table:parameter_biped}. To assess the performance of the proposed WB-DRC, two simulation studies are carried out. The first study involves applying an external force  along the $z$-axis to the robot's center of base link (see Fig. \ref{A1_gazebo}). In the second study, we simulate the robot walking while carrying an $8$ kg load, which accounts for up to $60$\% of its body weight. 

\begin{figure}[!ht]
	\centering
	\includegraphics[scale=0.027]{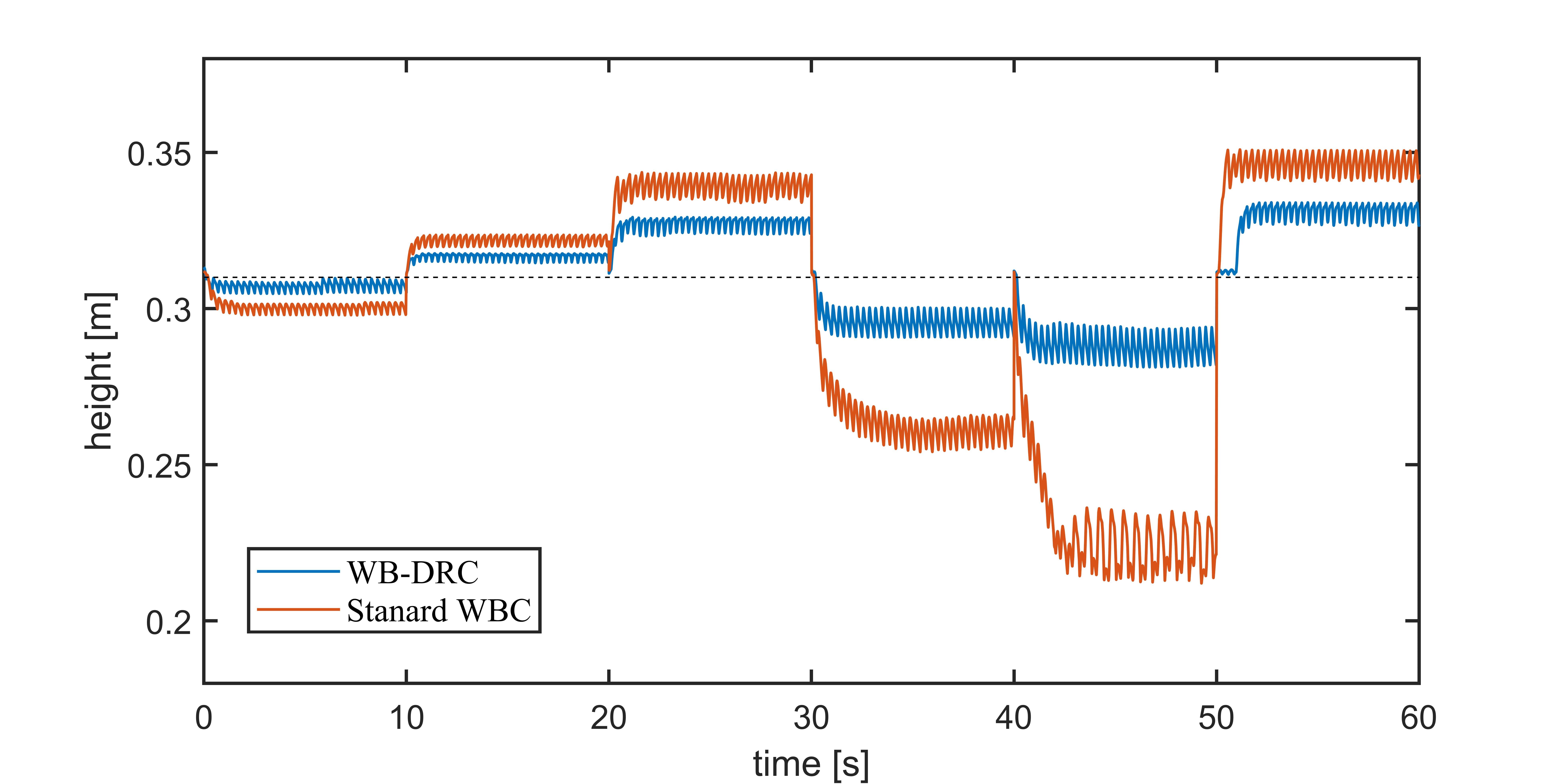}
	\caption{The height of the base link from the ground over time under the application of different external forces.}
	\label{fig_external_force}
\end{figure}

\begin{figure}[!ht]
	\centering
	\includegraphics[scale=0.027]{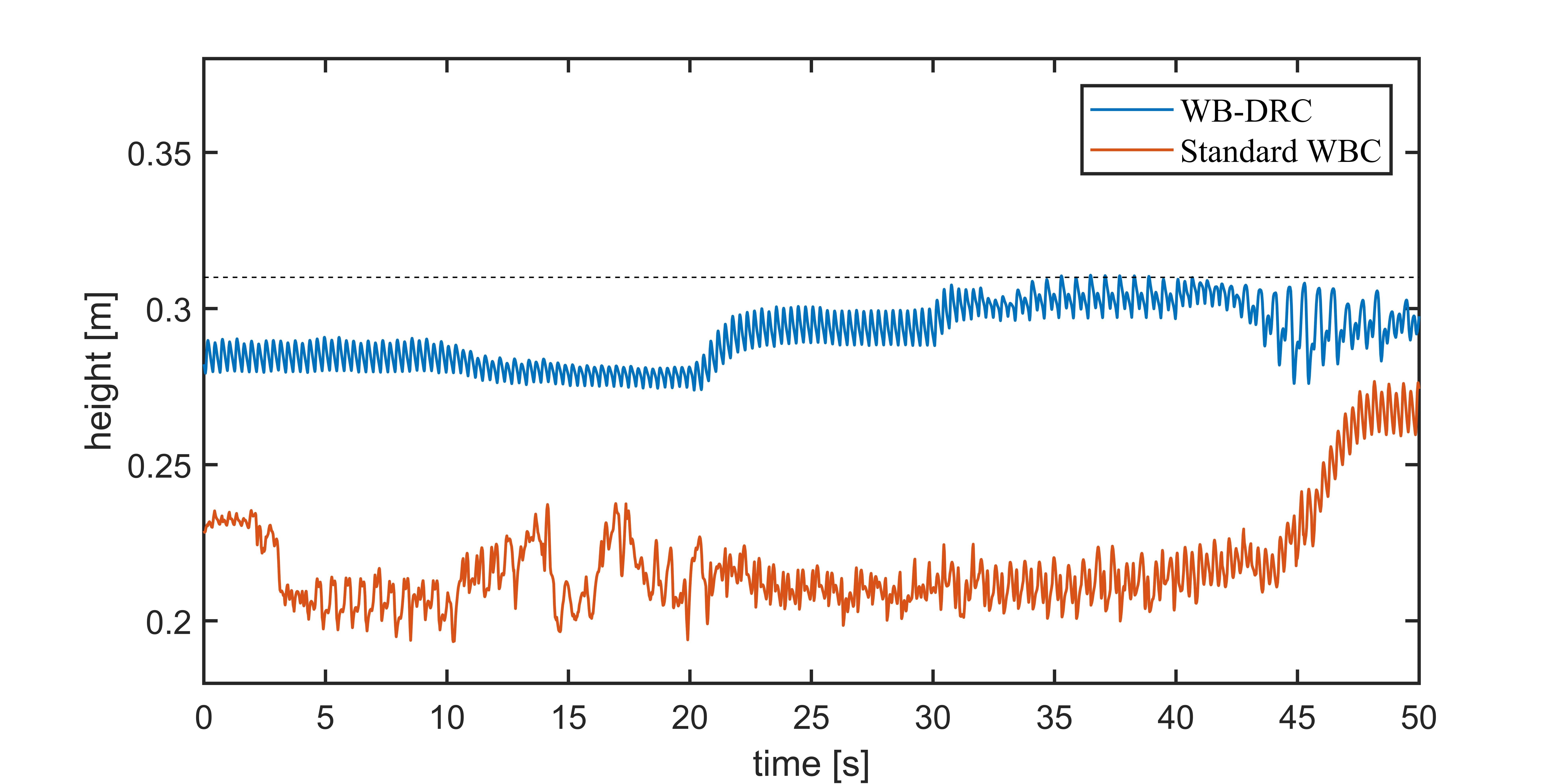}
	\caption{The height of the base link from the ground over time as the robot moves at different velocities with $8$kg load.}
	\label{fig_model_uncertainty}
\end{figure}

The desired height of the base link is $0.31$ m. Fig. \ref{fig_external_force} shows the height of the base link from the ground over time under the influence of various constant external forces. External forces of $-20$ N, $20$ N, $60$ N, $-60$ N, $-80$ N, and $80$ N are applied to the base link along the $z$-axis, each for a duration of 10 seconds.  The results demonstrate that the proposed WB-DRC compensates for these external disturbances better than  the standard WBC.
Fig. \ref{fig_model_uncertainty} illustrates the height of the base link over time as the robot with $8$ kg load walks at different velocities along the $x$-axis. The robot is commanded to move on the flat ground with velocities of -0.1 m/s, 0.1 m/s, 0.2 m/s, 0.4 m/s, and 0.6 m/s, each for a duration of 10 seconds. The results demonstrate that the WB-DRC  maintains the body height closer to the $0.31$ m, showing  its capability to effectively handle model uncertainties.

\section{Experiment Comparison}\label{sec:experiment}
We conduct two real-robot experiments on the Unitree A1 quadruped robot. In the first experiment, we simulate the motor malfunction by decreasing joint motor torque, and in the second experiment we compare  the performance of WB-DRC and standard WBC under the same  load. The control architecture is implemented on a  PC (Intel i7-13700KF, $3.4 $ GHz). The MPC operates at a frequency of $100$ Hz, {with a prediction horizon $T = 1$ s and the total number of steps $N = 30$,} while both the adaptive ESO and QP-WBC run at $1000$ Hz. The sliding window for the MAF is set to $5$. The controller parameters for the WB-DRC are provided in Table \ref{table:parameter_experiment}. The desired base link height is set to $0.3$ m.

\begin{figure}[!ht]  
	\centering
	\includegraphics[scale=0.0205]{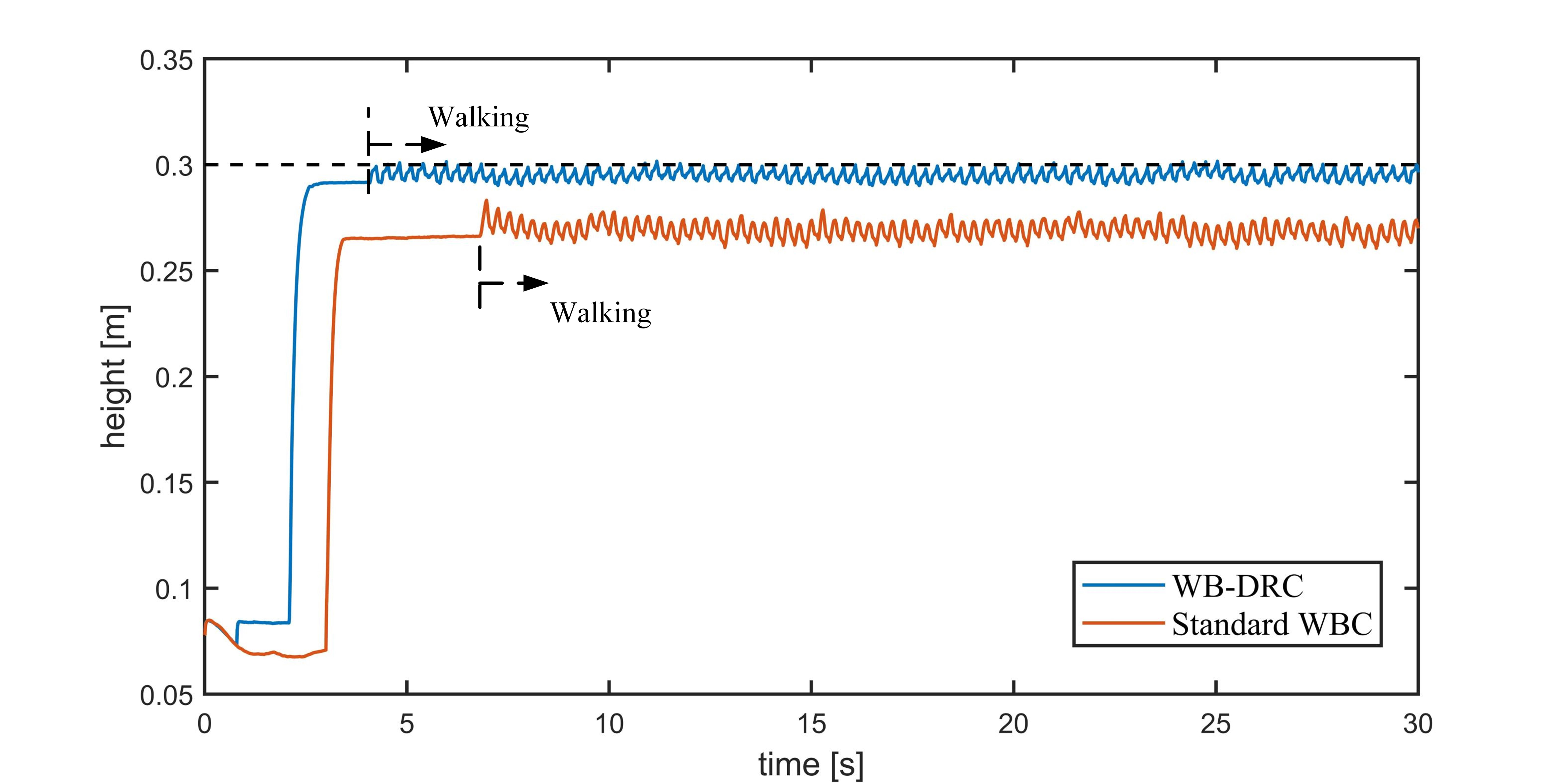}
	\caption{The base link height of the robot from standing to stepping in place.}
	\label{fig_exp_cut_torque}
\end{figure}

\begin{figure}[!ht]
	\centering
	\includegraphics[scale=0.028]{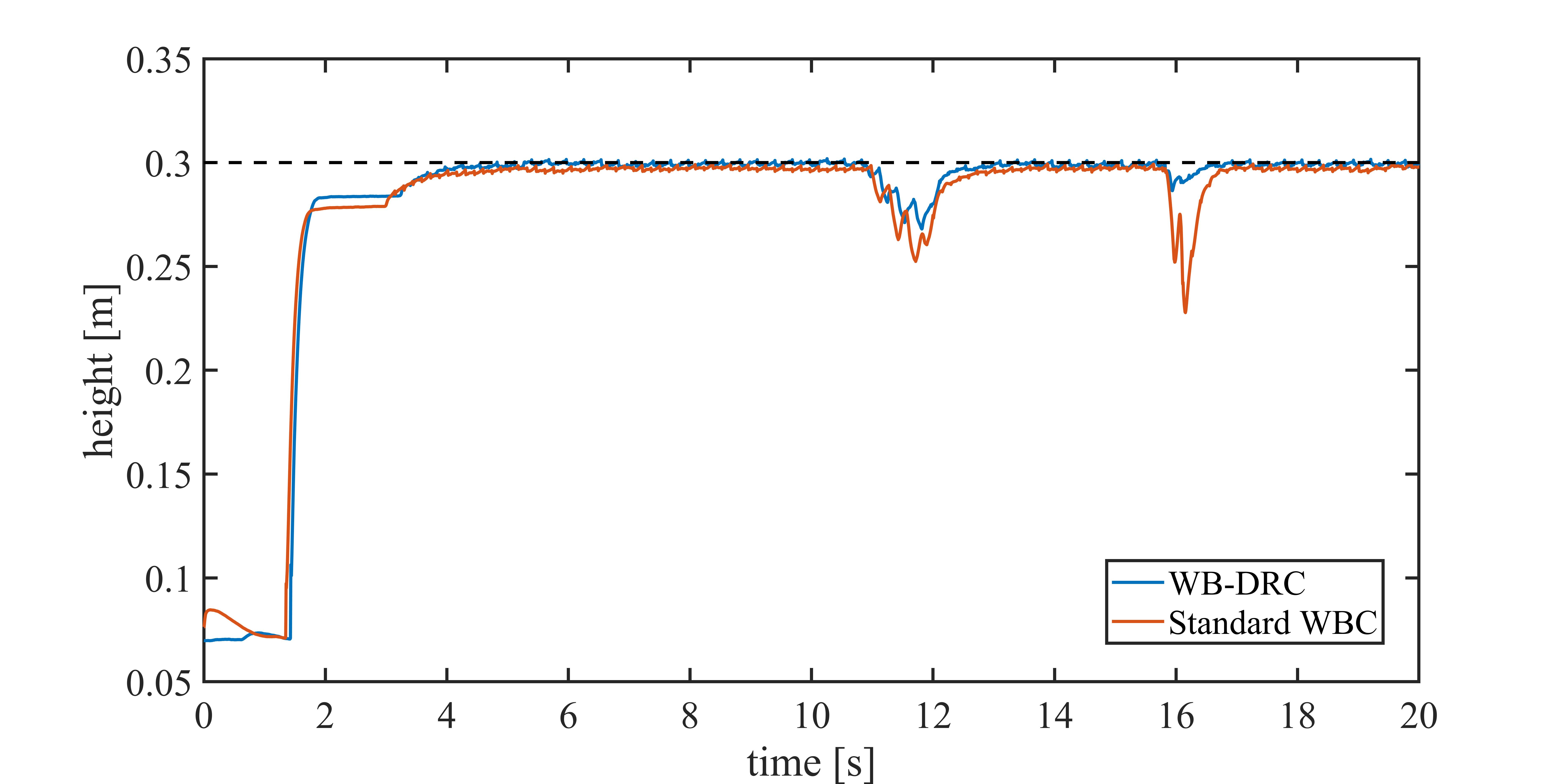}
	\caption{The base link height of the robot from standing to stepping in place.}
	\label{fig_exp_cut_torque_base}
\end{figure}

First, we intentionally reduce the output torque of the robot knee joint on the right back leg by $50$\%. The base link height of the robot, when the robot switches from standing to stepping in place, is shown in Fig. \ref{fig_exp_cut_torque} for  WB-DRC and standard WBC. For WB-DRC, the height trajectory of the base link closely follows the desired trajectories, outperforming the robot using the standard WBC. Second, we intentionally reduce both the output torque of the robot's knee joint and the output torque of the robot's hip joint on the right back leg by 60\% at 11s, with a duration of 1s, and by 90\% at 15.8s, with a duration of 0.2s. The base link height of the robot is shown in Fig. \ref{fig_exp_cut_torque_base} for  WB-DRC and standard WBC. The base link height of the robot using WB-DRC is less affected by short-term motor torque output failure than the robot using standard WBC. As shown in Figs. \ref{fig_exp_cut_torque} through \ref{fig_exp_cut_torque_base}, the robot using the  WB-DRC exhibits better fault tolerance and enhanced robustness compared to the robot using  the standard WBC.

\begin{table}[h]
	\begin{center}
		\caption{WB-DRC Setting}
		\begin{tabular}{l|l|l|l}
			\hline
			Parameter & Value &Parameter& Value\\
			\hline
			$\Gamma$ & $6000 \times I_{ 24 \times 24 }$&$Q_1$ & $500 \times I_{ 24 \times 24 }$\\
			\hline
			$\omega_0$ & $200$&$Q_2$ & $I_{ 18 \times 18 }$  \\
			\hline
			$\hat{\theta}(0)$ & $ \textbf{0}_{24 \times 1}$ & {$K_p$} & {$0 \times I_{12 \times 12}$} \\
			\hline
			$\theta_{\rm{max}}$ & $100 \times \textbf{1}_{24 \times 1}$ &{$K_d$} & {$3 \times I_{12 \times 12}$} \\
			\hline
			$\theta_{\rm{min}}$ & $-100 \times \textbf{1}_{24 \times 1}$ &&\\
			\hline
		\end{tabular}\label{table:parameter_experiment}
	\end{center}
\end{table}

%\begin{figure}[!ht]
%	\includegraphics[scale=0.028]{experiment_cut_torque_roll.jpg}
%	\caption{The roll angle of the base link when the output torque of the robot knee joint is reduced by $50$\%.}
%	\label{fig_exp_cut_torque_angle}
%\end{figure}

%\begin{figure}[!ht]
%	\includegraphics[scale=0.030]{experiment_cut_torque_pitch.jpg}
%	\caption{The pitch angle of the base link when the output torque of the robot knee joint is reduced by $50$\%.}
%	\label{fig_exp_cut_torque_pitch}
%\end{figure}

\begin{figure}[!ht]
	\includegraphics[scale=0.0215]{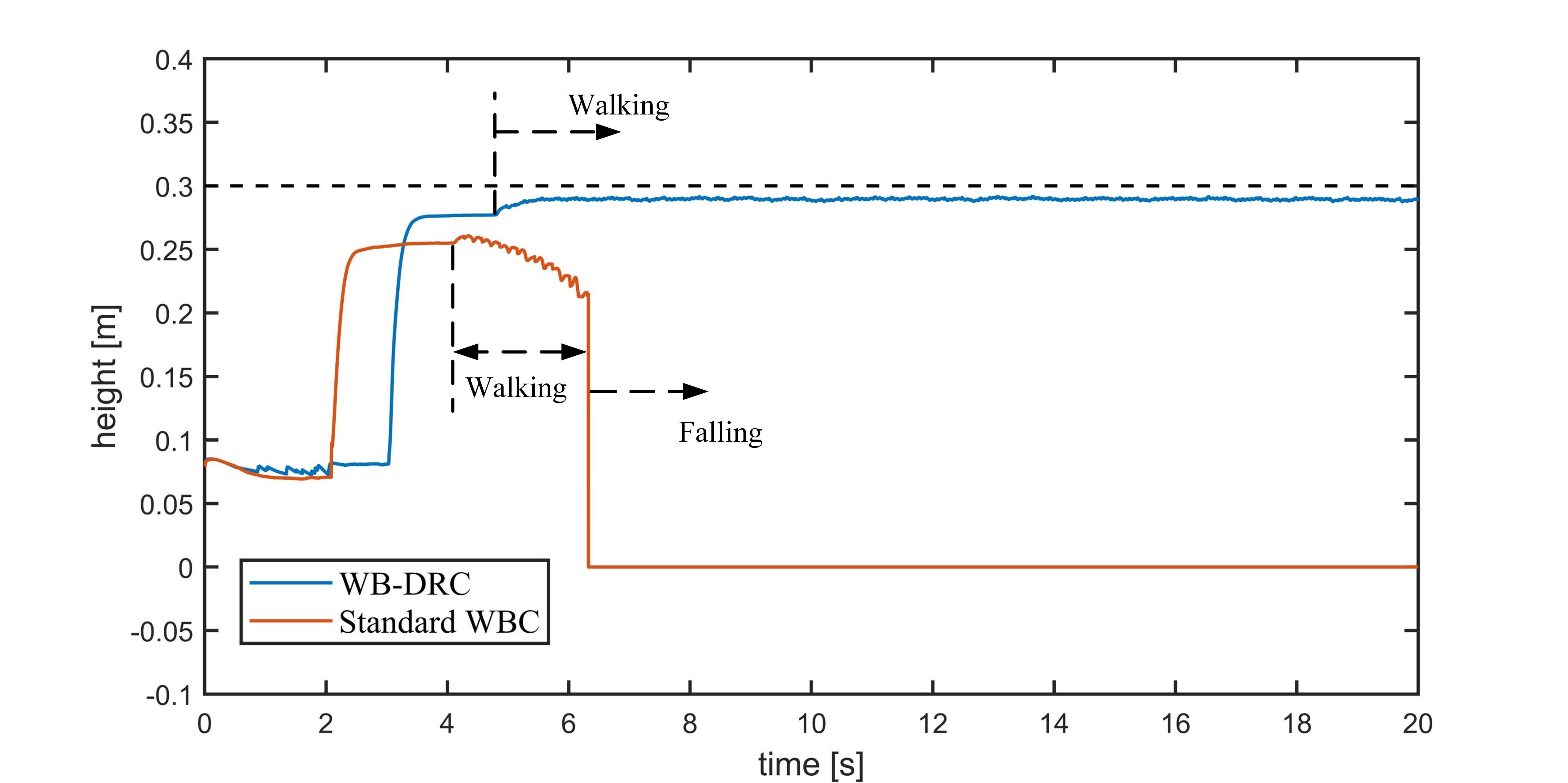}
	\caption{The height of the base link  when the robot using WB-DRC and standard WBC walks with 5 kg load.}
	\label{fig_exp_load_height}
\end{figure}

\begin{figure}[!ht]
	\centering
	\begin{minipage}{0.36\linewidth}
		\centering
		\includegraphics[width=0.9\linewidth]{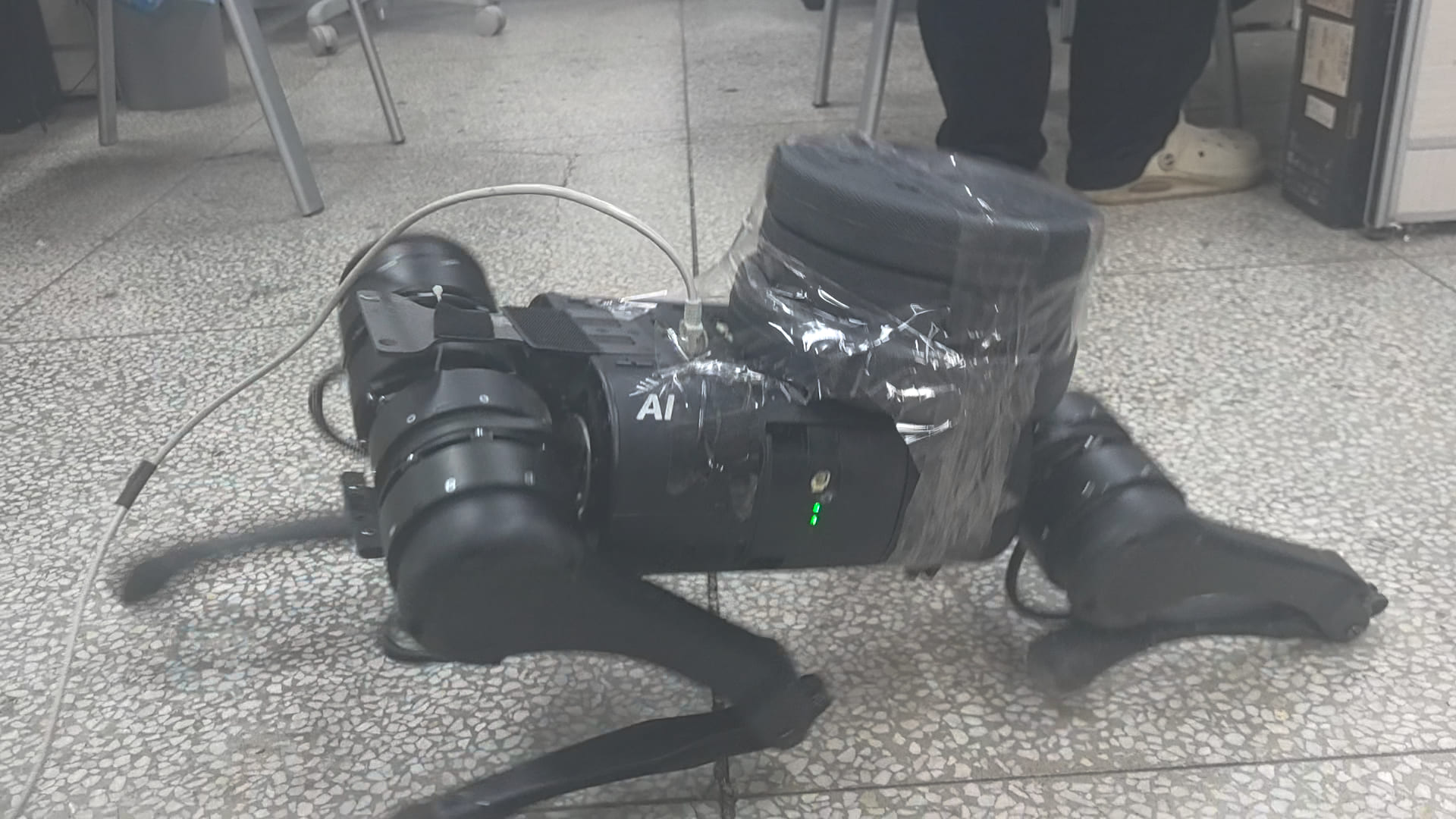}
	\end{minipage}
	\begin{minipage}{0.36\linewidth}
		\centering
		\includegraphics[width=0.9\linewidth]{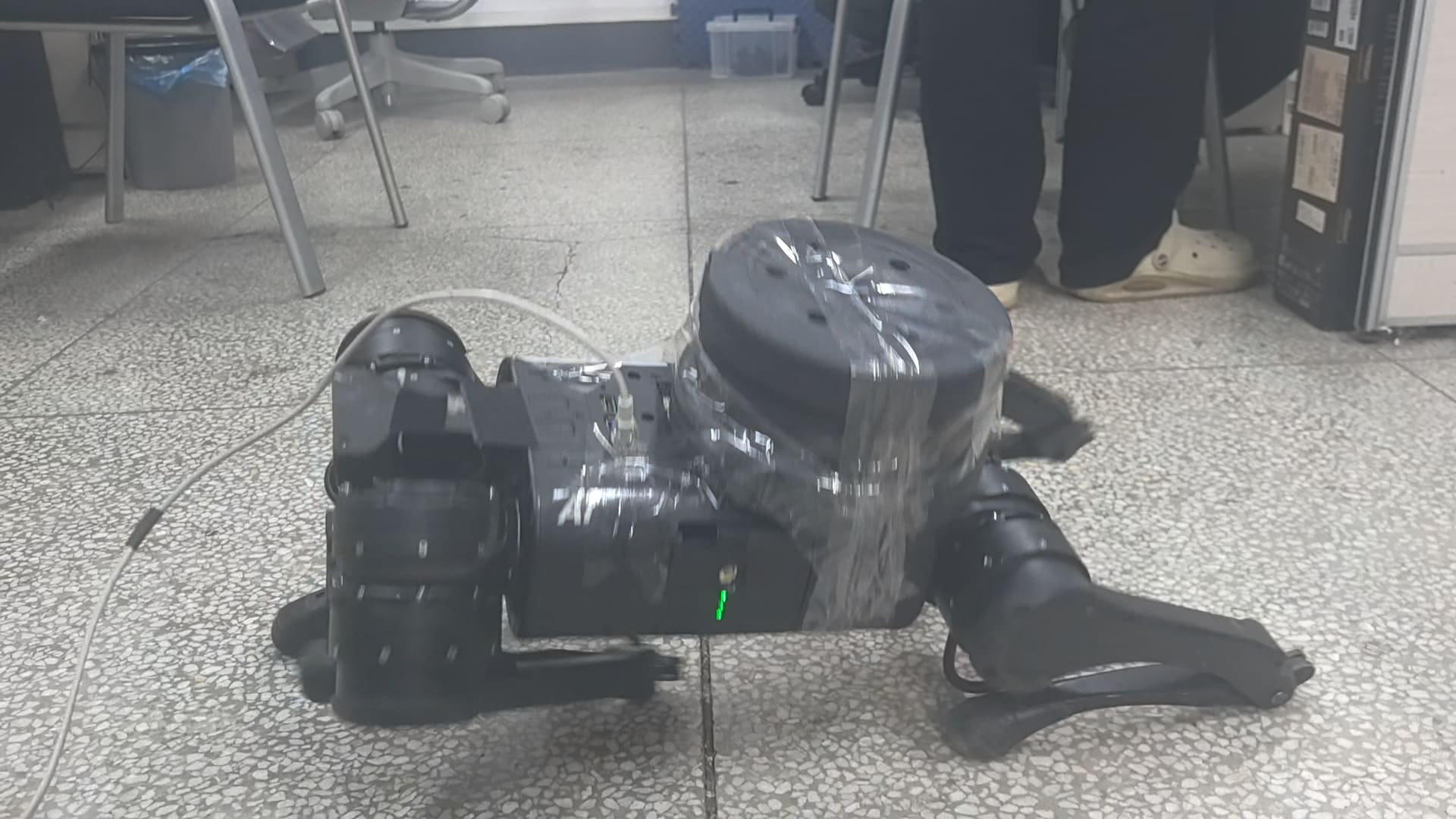}
	\end{minipage}
	\caption{The Unitree A1 robot using standard WBC falls at around 6 s.}
	\label{fig_biped_fall}
\end{figure}

\begin{figure}[!ht]
	\includegraphics[scale=0.028]{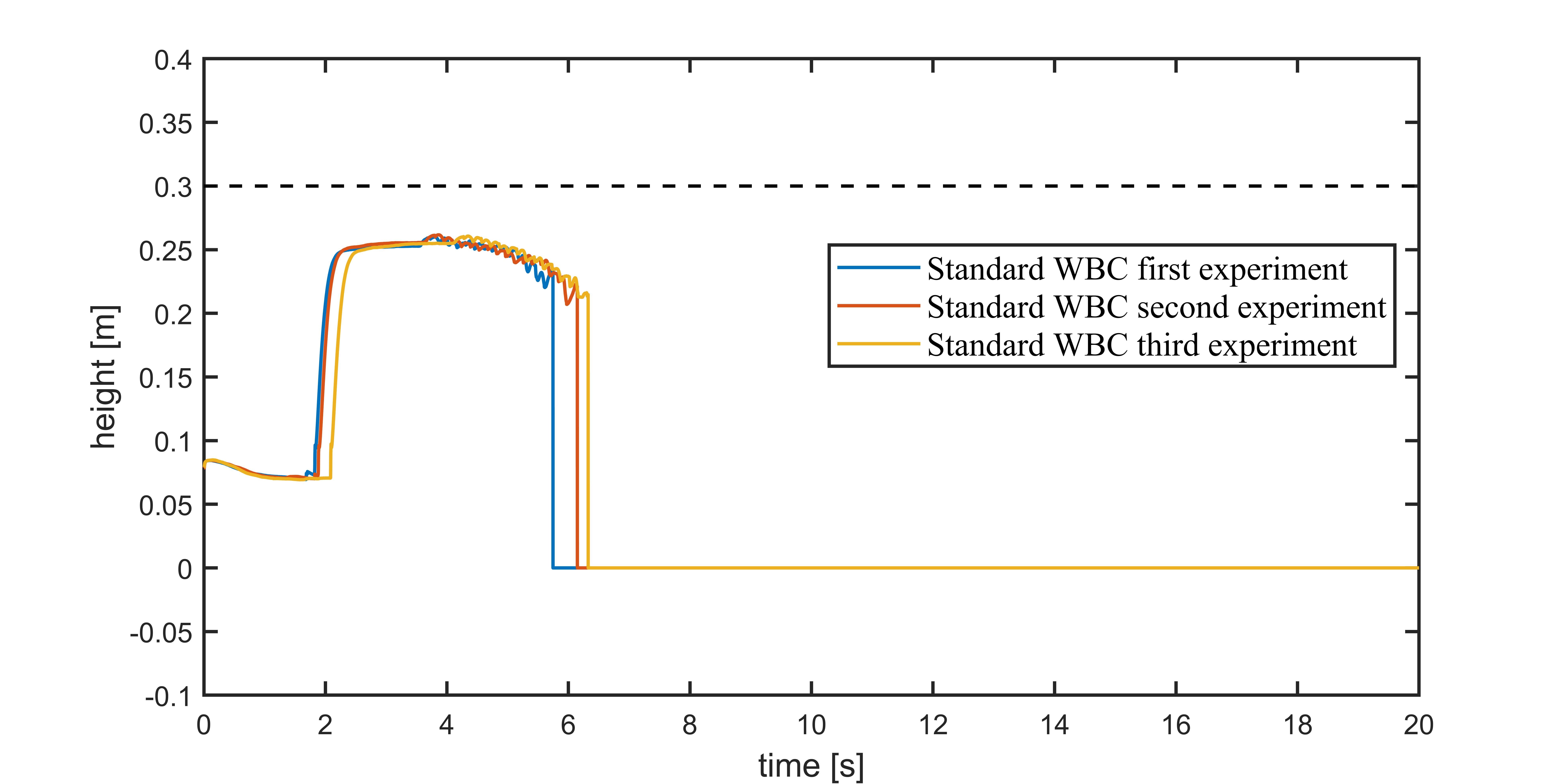}
	\caption{The height of the base link when the robot using standard WBC walks with $5$ kg load.}
	\label{fig_exp_standardWBC}
\end{figure}

To further evaluate the performance of WB-DRC, {we conducted the experiment where the robot carries a 5 kg load.} Fig. \ref{fig_exp_load_height} illustrates the height of the robot's base link under both WB-DRC and standard WBC. The robot using the standard WBC falls at around 6 s, as shown in Fig. \ref{fig_biped_fall}, while the robot with the proposed WB-DRC is able to walk continuously for the full 20 seconds. We performed three consecutive tests on the robot carrying a 5 kg load using the standard WBC. The results, shown in Fig. \ref{fig_exp_standardWBC}, indicate that the standard WBC is unable to effectively handle the robot's 5 kg load, whereas the WB-DRC performs successfully. These experiments demonstrate that the WB-DRC offers significant advantages in terms of robustness and stability.

%\begin{figure}[!ht]
%	\centering
%	\includegraphics[scale=0.065]{experiment_platform.jpg}
%	\caption{The Unitree A1 robot with 5 kg load.}
%	\label{fig_exp_experiment_platform}
%\end{figure}

\begin{figure}[!ht]
	\centering
	\includegraphics[scale=0.25]{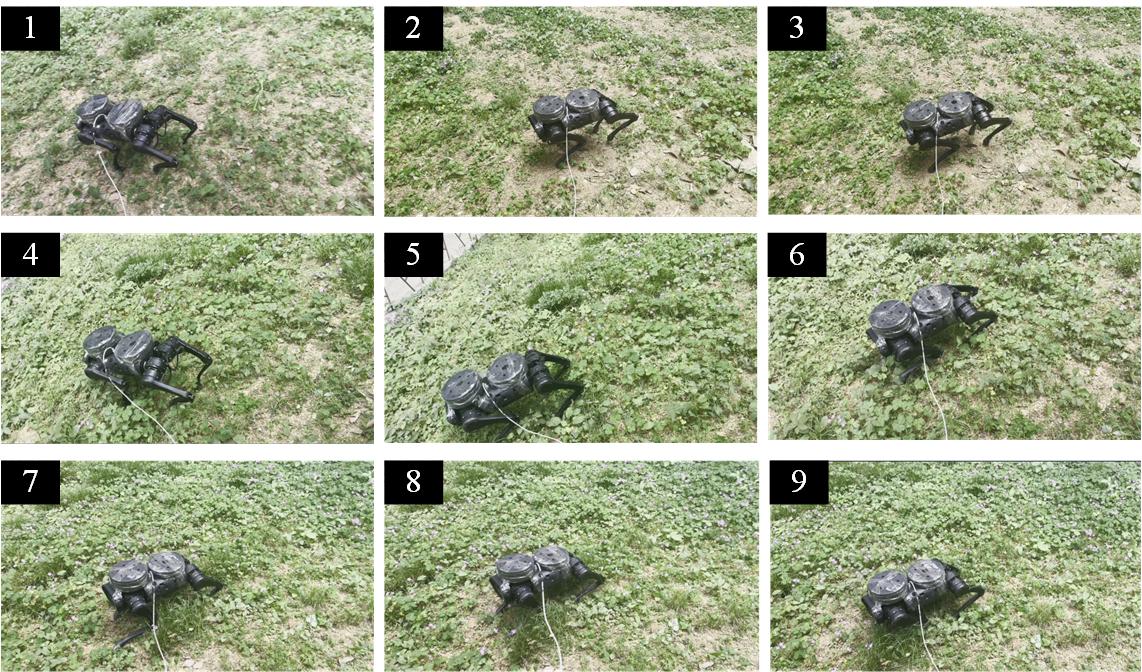}
	\caption{The robot moves on tough terrain while carrying a 5 kg load.}
	\label{fig_exp_tough_load}
\end{figure}

{To demonstrate the capability of the control framework in handling model uncertainties and external disturbances on rough terrain, we conduct a real robot experiment on challenging terrain. As shown in Fig. \ref{fig_exp_tough_load}, the robot successfully navigates the rough terrain at significant velocity while carrying a 5 kg load, highlighting the effectiveness of the proposed framework in enabling the robot to traverse difficult environments.}

\section{Conclusions}\label{sec:conclusions}
%In this letter, we present a novel control framework for legged robots that enhances their self-perception capabilities and robustness against external disturbances and model uncertainties. The proposed disturbance estimator, which integrates adaptive control with ESO, effectively estimates both external disturbances and model uncertainties, and compensates for them within the whole-body control framework. The introduction of the WB-DRC significantly improves the robot's ability to handle uncertainty, surpassing traditional whole-body control approaches. Through simulation results for both biped and quadruped robots, as well as extensive experimental validation on the Unitree A1 quadruped robot, we have demonstrated the effectiveness, robustness, and stability of the proposed framework. This research offers a promising solution for enhancing the performance of legged robots in dynamic environments and highlights their potential to adapt to unpredictable conditions.

{In this letter, we present a novel control framework for legged robots that enhances self-perception and robustness against external disturbances and model uncertainties. The proposed disturbance estimator, which combines adaptive control with an ESO, effectively estimates and compensates for disturbances and uncertainties within the whole-body control framework. The WB-DRC improves the robot's ability to handle various uncertainties, particularly those related to joint motor output torque inaccuracies, while also expanding the application range, making it suitable not only for quadruped robots but also for biped humanoid robots, surpassing traditional methods. Simulation results for both biped and quadruped robots, along with experiments on the Unitree A1 quadruped, demonstrate the framework's effectiveness, robustness, and stability. This research offers a promising solution for enhancing the performance of legged robots in dynamic environments.}

\bibliographystyle{IEEEtran}
\bibliography{IEEEabrv,IEEEexample}
\end{document}